\PassOptionsToPackage{table}{xcolor} %
\documentclass{article}

\PassOptionsToPackage{numbers,sort&compress}{natbib}
\usepackage[main, final]{neurips_2026}
\usepackage[utf8]{inputenc}
\usepackage[T1]{fontenc}
\usepackage{hyperref}
\hypersetup{pdftitle={Closing the Loop: Co-Evolving EM for Irregular Time Series Generation in Lifted Representations},pdfauthor={Idan Arbiv, Gal Fadlon, Omri Azencot}}
\usepackage{url}
\usepackage{float}
\usepackage{booktabs}
\usepackage{amsfonts}
\usepackage{xcolor}
\usepackage{nicefrac}
\usepackage{microtype}
\usepackage[table]{xcolor}
\usepackage{graphicx}
\graphicspath{{figs/}}
\usepackage{overpic}
\usepackage{multirow}
\usepackage{amsmath}
\usepackage{amssymb}
\usepackage{mathtools}
\usepackage{amsthm}
\usepackage{algorithm}
\usepackage{algpseudocode}

\usepackage[capitalize,noabbrev]{cleveref}
\newcounter{appmech}[subsection]

\usepackage{colortbl}
\usepackage{pifont}

\providecommand{\xmark}{\ding{55}}
\theoremstyle{plain}
\newtheorem{theorem}{Theorem}[section]
\newtheorem{proposition}[theorem]{Proposition}

\newtheorem{corollary}[theorem]{Corollary}
\theoremstyle{definition}

\theoremstyle{remark}
\newtheorem{remark}[theorem]{Remark}
\usepackage{wrapfig}
\usepackage{subcaption}
\usepackage{tikz}
\usepackage{enumitem}
\usetikzlibrary{arrows.meta,positioning,shapes.geometric,calc,decorations.pathmorphing}
\newenvironment{restatedprop}[1]{\par\medskip\noindent\textbf{Proposition~\ref{#1}} \textbf{(restated)}\textbf{.}\ \itshape}{\par\medskip}
\newcommand{\tsimg}{\mathcal{T}}
\newcommand{\imgts}{\mathcal{T}^{-1}}
\newcommand{\pslr}{\textsc{pslr}}
\newcommand{\lmmps}{L\mbox{-}\textsc{mmps}}
\newcommand{\ltmpd}{L\mbox{-}\textsc{tmpd}}
\newcommand{\Nts}{N_{\mathrm{ts}}}
\newcommand{\Nimg}{N_{\mathrm{img}}}
\newcommand{\Nobs}{N_{\mathrm{obs}}}
\newcommand{\Ats}{A_{\mathrm{ts}}}
\newcommand{\Aimg}{A_{\mathrm{img}}}
\newcommand{\yts}{y_{\mathrm{ts}}}
\newcommand{\xts}{x_{\mathrm{ts}}}
\newcommand{\ximg}{x_{\mathrm{img}}}

\title{Closing the Loop: Co-Evolving EM for Irregular \\ Time Series Generation in Lifted Representations}

\author{%
  Idan Arbiv\thanks{Equal contribution.} \quad Gal Fadlon\footnotemark[1] \quad Omri Azencot \\
  Department of Computer Science \\
  Ben-Gurion University of The Negev \\
  \texttt{\{arbivid, galfad\}@post.bgu.ac.il} \\
  \texttt{azencot@bgu.ac.il} \\
}

\begin{document}

\maketitle

\begin{abstract}

Regular time-series generation from irregular time series is typically reduced to one-shot imputation: complete each partial sequence once, freeze the completions, and train a generator on the resulting surrogate dataset. We show that this open-loop strategy is structurally flawed. Point imputers collapse ambiguous missing regions to conditional averages, while stochastic imputers become stale as the generator evolves; the binding constraint is therefore iteration, not imputer design. We introduce a co-evolving Monte Carlo EM framework that closes the imputer--generator loop. Each E-step samples missing values from the posterior of the current generator, and each M-step retrains the generator on these completions, preserving unconditional generator training. Since our diffusion prior operates in a lifted 2D representation while observations live in time-series space, we further introduce Posterior Sampling for Lifted Representations (PSLR), which enforces operator-consistent conditioning, uncertainty preservation, and manifold consistency during the E-step. Across ten datasets and four corruption regimes, our method achieves state-of-the-art performance, improving over the strongest baselines by 59\% in discriminative score, 15\% in predictive score, 73\% in Context-FID, and 71\% in feature-correlation error. Ablations show that our approach reduces average discriminative score by 73\% over single-step baselines, closing most of the remaining gap to the clean-data oracle. Code is available at \url{https://github.com/azencot-group/PSLR}.

\end{abstract}

\section{Introduction}
\label{sec:intro}
\vspace{-1mm}

Time series underpin forecasting, anomaly detection, and pattern analysis across healthcare, finance, and the sciences~\citep{han2019review,ismail2019deep,lim2021time}, and \emph{generating} synthetic series is increasingly valuable for privacy-preserving proxies, stress testing, and dataset balancing~\citep{brophy2023generative,nochumsohn2025data,gonen2025time}. In practice, however, real-world series are often \emph{irregular}: observations arrive at uneven times and values are missing due to sensor failures, inconsistent sampling, or interruptions in monitoring~\citep{kidger2020ncde,ren2024learning}. We study generative learning under such corruption: each sequence is only partially observed, yet the goal is to learn the distribution of the underlying time series before corruption.

Prior methods reduce irregular generation to a missing-value problem and handle it \emph{open-loop}: the mechanism that handles missingness is applied once at the start and never updated. The dominant version is a \emph{commit-and-forget} pipeline in which a pretrained imputer is run once over each irregular sequence and its completions are then frozen for generator training~\citep{jeon2022gtgan, naiman2024kovae, fadlon2025imageni2r}. These imputers are typically trained to produce \emph{point estimates} of the missing values, often by minimizing mean-squared error. However, missing values are inherently ambiguous: the same partial sequence can be completed in multiple valid ways. A point imputer resolves this ambiguity by selecting a single completion for each sequence, typically corresponding to a summary of the conditional distribution (e.g., the mean under squared-error training~\citep{bishop2006pattern,murphy2012machine}). Applied across the dataset, this systematically reduces variability, so a generator trained on these completions learns a narrower distribution than the true one (see \cref{fig:cond-mean-trap}), leading to degraded generative quality.

One might hope that a better imputer would fix this. It would not, because the issue is structural: existing methods treat imputation as a one-shot preprocessing step, independent of the generator it is meant to support. As a result, the completions used for training are fixed before learning begins and cannot adapt as the generator improves. Yet the notion of a “plausible” completion depends on the generator itself: an early-stage generator and a well-trained one disagree about which completions are consistent with the data. A fixed imputer cannot track this evolving notion, and therefore provides the generator with systematically mismatched training data, even when its reconstruction error is low (\cref{tab:illustration-mse-trap}). This suggests that irregular time-series generation requires closed-loop training, rather than one-shot imputation. \textbf{Namely, the binding constraint is iteration: imputations must be refined repeatedly using feedback from the generator they are used to train.}

To realise this iterative feedback, we close the imputer–generator loop using a co-evolving Monte Carlo EM (MCEM) framework~\citep{dempster1977em,wei1990mcem,neal1998em}. At each iteration, the E-step samples missing values from the generator’s current posterior (rather than producing point estimates), and the M-step retrains the generator on these completed sequences; the two are alternated for a small number of iterations. MCEM is well suited to this setting for three reasons (\cref{sec:why-close}). First, in its idealized form (exact E-step) it guarantees monotone improvement of the data likelihood, in contrast to alternative iterative schemes that lack such guarantees. Second, once the E-step produces completed data, the M-step reduces to standard training on fully observed sequences, without masking or auxiliary components~\citep{fadlon2025imageni2r}. Third, by sampling from the model’s own posterior, the E-step produces completions that remain consistent with the current generator, avoiding the mismatch introduced by fixed, pretrained imputers. We instantiate this framework with a diffusion-based generator for time series built on a vision backbone~\citep{naiman2024imagentime, gonen2025time, fadlon2025imageni2r}. By lifting sequences into 2D representations, we can bypass the limitations of native 1D models in capturing long-range dependencies and directly leverage the rapid advancements and high-capacity architectures of the image diffusion community. This approach lifts each sequence into a $2$D representation (e.g., via delay embeddings or spectrogram-style transforms) and applies image-space diffusion. After training, this yields an unconditional generator that can be sampled directly using standard diffusion methods.

While the generator is trained as an unconditional model, the E-step of MCEM requires sampling complete series that are consistent with partial observations. This amounts to drawing from the posterior of the current generator conditioned on the observed data. Rather than training a separate conditional model alongside the unconditional one, we instead perform posterior sampling~\citep{boys2023tmpd, rozet2024diem}, combining the learned prior with an observation model that enforces consistency with the data. In our setting, however, this is non-trivial: the generator operates in a lifted $2$D representation, while observations are defined in the original time-series space. As a result, conditioning imposes constraints in the time domain that must be enforced through an operator acting on the lifted representation. Existing posterior samplers are derived for the single-space setting where prior and observations coincide, and when applied here they either fail to incorporate the observations or perform poorly (\cref{tab:samplers-motivation}). To address this, we introduce \emph{Posterior Sampling for Lifted Representations} (\pslr{}), a plug-and-play approach for posterior sampling under cross-space constraints. We first identify the conditions a sampler must satisfy to serve as a valid E-step under a lifted prior (\cref{subsec:pslr}), and then show how to enforce them in practice. Adapting existing samplers using \pslr{} yields variants that operate correctly in the lifted setting. Without these conditions, the E-step solves a different inverse problem and the EM loop converges to a biased solution (App.~\ref{app:mmps-fails}); with them, it reliably recovers the correct generator (\cref{tab:ablation}).  Our key contributions are summarized as follows:

\begin{enumerate}
\setlength{\itemsep}{2pt}

\item \textbf{Iteration is the binding constraint.} We prove that training on completions from any frozen imputer leaves a bias that closed-loop EM removes (Proposition~\ref{prop:frozen-bias}, \cref{sec:why-close}), and show empirically across a $10$-dataset benchmark that closing the loop yields consistent improvements over all published open-loop baselines. Our results demonstrate that iteration, not imputer design or architectural sophistication, is the primary driver of generative fidelity under missingness.

\item \textbf{A closed-loop framework for irregular generation.} We introduce a co-evolving Monte Carlo EM (MCEM) framework that alternates between sampling missing values and retraining the generator (\cref{subsec:closed-loop}). The resulting model is an unconditional generator trained on completed data, with conditioning handled via posterior sampling. To enable this in the lifted setting (\cref{subsec:lifted-posterior}), we develop \pslr{}, a plug-and-play approach for posterior sampling under cross-space constraints, with three structural conditions (C1--C3) that any valid dual-space E-step sampler must satisfy (\cref{subsec:pslr}).

\item \textbf{State-of-the-art (SOTA) performance.} Our method sets a new SOTA across the standard ten-dataset benchmark, recovering near-oracle performance with rapid convergence (\cref{sec:results}). We further achieve SOTA on three additional corruption types beyond the standard timestamp-level setting - per-cell asynchronous-channel missingness, contiguous block gaps, and continuous-time fractional resampling (\cref{fig:corruption-types}), each of which breaks the assumptions of standard one-shot imputation pipelines (\cref{tab:avg-benchmarks}).

\end{enumerate}

\vspace{-2mm}
\section{Related Work}
\label{sec:related}
\vspace{-1mm}

\paragraph{Irregular time-series generation.}
Early work adapted regular-grid generators (TimeGAN, RCGAN, C-RNN-GAN, T-/P-Forcing) to the irregular setting via $\Delta t$ concatenation or learned exponential decay (GRU-$\Delta t$, GRU-D)~\citep{che2018recurrent}; GT-GAN~\citep{jeon2022gtgan} introduced these as its $\Delta t$/-D baseline suite. The dedicated line then proceeds GT-GAN $\rightarrow$ KoVAE~\citep{naiman2024kovae} $\rightarrow$ ImagenI2R~\citep{fadlon2025imageni2r}, all sharing a two-stage \emph{commit-and-forget} template: GT-GAN encodes the irregular series with a neural-CDE~\citep{kidger2020ncde} and decodes with a GAN; KoVAE keeps the NCDE preprocessing but swaps the GAN for a Koopman-structured VAE that imposes (approximately) linear latent dynamics; ImagenI2R first imputes with a Time Series Transformer~\citep{zerveas2021tst} and then trains a vision diffusion on ImagenTime~\citep{naiman2024imagentime} delay-embedding images with a masked loss on completed pixels. All three are \emph{open-loop}: the one-shot completion (NCDE preprocessing or masked transformer) is produced once and frozen, so the imputer cannot benefit from a better generator and the generator cannot correct a mis-specified imputer.

\paragraph{Closing the imputer--generator loop.}
The Expectation--Maximization (EM) algorithm~\citep{dempster1977em} is the canonical framework for learning with latent or partially-observed variables and has underpinned the multiple-imputation literature for decades~\citep{rubin1987multiple}. Three properties make it the right machinery here: (i)~a monotone observed-data likelihood guarantee for the idealized loop with an exact E-step, (ii)~an M-step that reduces to standard training on completed data, and (iii)~an E-step that samples from the model's own posterior. Natural alternatives sacrifice at least one. \emph{Self-training} alternates filling missing values with the generator's own predictions and retraining on those completions; this keeps~(ii), but a single prediction is not a posterior sample~(iii), and the iteration carries no likelihood guarantee~(i). \emph{Variational EM}~\citep{neal1998em,kingma2014vae} replaces sampling with a second network that proposes completions, trained alongside the generator; because this network only approximates rather than samples the posterior, all three guarantees are weakened, and no clean instantiation exists under a diffusion prior. We adopt Monte-Carlo EM~\citep{wei1990mcem} with incremental M-steps~\citep{neal1998em}; \cref{sec:why-close} develops the guarantees of our approach.

\paragraph{Diffusion posterior sampling.} Posterior sampling provides a way to guide an unconditional diffusion model using observations at inference time, without training a separate conditional model. Concretely, given a prior $p_\theta(x)$ and measurements, these methods generate samples that both follow the prior and satisfy the observations by iteratively correcting the reverse-diffusion trajectory. There are multiple samplers derived under a \emph{single-space} assumption (observation operator in the model's native space), differing in how they approximate conditioning: DPS~\citep{chung2023dps} uses a point estimate, $\Pi$GDM~\citep{song2023pigdm} an isotropic Gaussian, DiffPIR~\citep{zhu2023diffpir} a proximal update, TMPD~\citep{boys2023tmpd} a row-sum diagonal Tweedie covariance, and MMPS~\citep{rozet2024diem} the full Jacobian Tweedie covariance via conjugate gradient. However, as discussed in \cref{sec:background}, our setting is \emph{dual-space}: observations live in the time-series domain, while the diffusion prior is defined over lifted image representations. In this setting, only TMPD and MMPS remain structurally compatible: the first three either collapse posterior diversity, lose the lift’s correlation structure, or admit no closed form when composed with the lift.

\vspace{-2mm}
\section{Background}
\label{sec:background}
\vspace{-1mm}

\begin{figure}[t]
\centering
\includegraphics[width=\textwidth]
{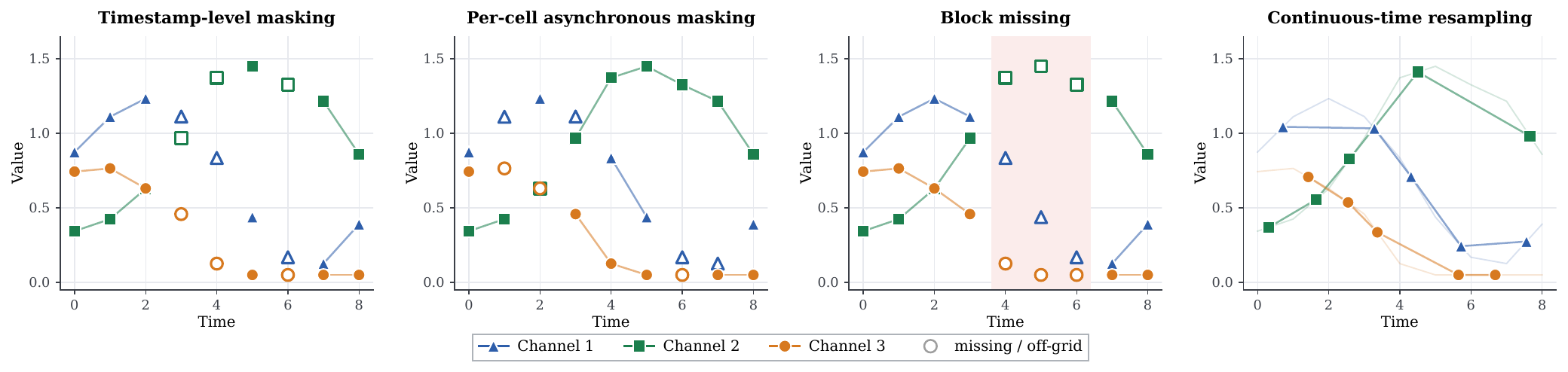}
\vspace{-2mm}
\caption{\textbf{Four corruption regimes covered by a single operator $\Ats$.} Three-channel signals; filled/hollow markers denote observed/missing entries. \emph{Timestamp-level}: all channels drop together; \emph{Per-cell}: each $(t,c)$ drops independently; \emph{Block}: a contiguous segment is unobserved; \emph{Continuous-time resampling}: each channel is sampled on its own fractional grid (via linear blending).}
\label{fig:corruption-types}
\vspace{-2mm}
\end{figure}

\paragraph{Problem statement.}
We observe a dataset of irregular time series $\{ (\yts^{(i)}, \Ats^{(i)}) \}_{i=1}^N$, where each sequence is a partially observed version of an underlying complete, regularly-sampled time series $\xts^{(i)} \in \mathbb{R}^{\Nts \times F}$. Here, $\Nts$ denotes the number of time steps and $F$ denotes the number of features per time step. We model the observation process by an operator $\Ats^{(i)}$ defined as:
\[
\Ats^{(i)} : \mathbb{R}^{\Nts \times F} \to \mathbb{R}^{\Nobs^{(i)} \times F} \ ,
\qquad
\yts^{(i)} = \Ats^{(i)} (\xts^{(i)}) \ .
\]
This notation covers the four corruption regimes we evaluate on (\cref{fig:corruption-types}): timestamp-level masking (the standard benchmark), per-cell asynchronous masking, contiguous block gaps, and continuous-time fractional resampling. The goal is to learn the marginal distribution $p(\xts)$ and generate complete, regularly-sampled time series.

\paragraph{Preliminaries.} We build on diffusion-based generative modeling, where a denoiser $d_\theta(x, \sigma)$ is trained to predict clean data from noisy inputs at varying noise levels $\sigma$ using a weighted denoising objective~\citep{karras2022edm}. After training, new samples are generated by iteratively denoising Gaussian noise. To apply this framework to time series, we follow prior work and lift each sequence into a 2D representation via a linear map $\tsimg$, with inverse $\imgts$~\citep{naiman2024imagentime}. The generator therefore models a distribution over lifted image variables $\ximg = \tsimg(\xts)$. Finally, we use posterior sampling to generate samples consistent with observations. Given a prior $p_\theta(x)$ and measurements $y = Gx$, posterior samplers iteratively correct samples so that they both follow the prior and satisfy the measurement constraints.

\vspace{-2mm}
\section{Method}
\label{sec:method}
\vspace{-1mm}

We cast irregular time-series generation as a closed-loop training problem. Existing methods operate open-loop: they complete the data once using a fixed imputer and then train a generator on those frozen completions. We instead use Monte Carlo EM. During training, the E-step performs posterior sampling under the current unconditional generator to draw complete series consistent with the observations, and the M-step retrains the generator on these sampled completions.

This formulation imposes a key constraint: the E-step must generate samples consistent with the observations, even though the generator itself is unconditional. Posterior sampling provides the bridge: it conditions the current unconditional prior on the observed data without introducing a separate conditional model. In our setting, however, this posterior inference is not standard. The generator operates in a lifted image space, while the observations live in the original time-series space. Thus the sampler must enforce time-series constraints on samples represented as images.

We address this mismatch with \pslr{}, a posterior-sampling recipe for lifted representations. We first identify three requirements any E-step sampler must satisfy in this dual-space setting, then instantiate them with MMPS. The rest of this section justifies the need for closed-loop training, formalizes the MCEM objective, and derives the lifted E-step used in our algorithm.

\paragraph{Scope of the theory.} The lift and the theory play different roles. Lifting series to images is a modeling choice: it lets us reuse strong vision diffusion backbones and scale to very long sequences~\citep{naiman2024imagentime}. The theory answers two questions. First, why iterate: training on completions from any frozen imputer leaves a bias, and iterating with the current generator removes it (Proposition~\ref{prop:frozen-bias}). Second, how to sample across the two spaces: we show where single-space samplers go wrong in the lifted setting (App.~\ref{app:mmps-fails}), and each of our corrections C1--C3 rests on a formal statement. They hold for any linear lift, and swapping delay embedding for STFT changes results by about $1\%$ (App.~\ref{app:stft}).

\vspace{-1mm}
\subsection{Why closing the loop matters}
\label{sec:why-close}
\vspace{-1mm}

\begin{figure}[t]
\centering
\includegraphics[width=0.99\textwidth]{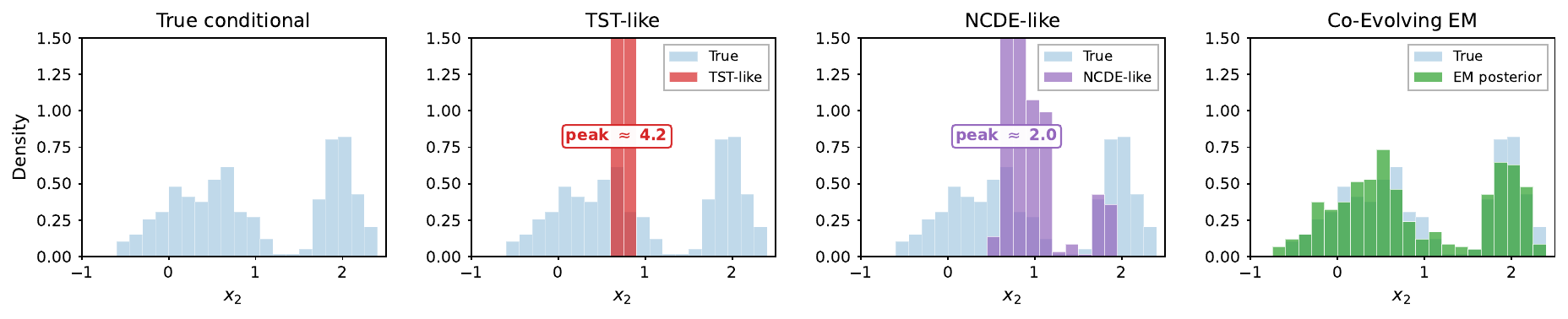}
\vspace{-3mm}
\caption{\textbf{Mean imputation vs.\ posterior sampling on a 2D illustration.} With a bimodal conditional $p(x_2 \mid x_1 \!\approx\! 0)$, deterministic TST-like~\citep{zerveas2021tst,fadlon2025imageni2r} and NCDE-like~\citep{kidger2020ncde,jeon2022gtgan,naiman2024kovae} imputers collapse missing values to the conditional mean, placing mass between the true modes. Our co-evolving EM samples from the current generator posterior and recovers both modes. Full metrics in App.~\ref{app:illustration} (\cref{tab:illustration-mse-trap}).}
\label{fig:cond-mean-trap}
\vspace{-2mm}
\end{figure}

Open-loop pipelines train the generator on completions produced by a pretrained and frozen imputer. When missing values are ambiguous, an MSE-trained imputer resolves this ambiguity by predicting the conditional mean (\cref{fig:cond-mean-trap}), which can lie between modes in regions where the data has no support. As a result, the dataset is systematically distorted, and the generator is trained on a narrowed version of the true distribution~\citep{rubin1987multiple} (see App.~\ref{app:cond-mean-trap}--\ref{app:var-collapse}).

These failures are structural, not artifacts of a weak imputer. Training only on observed entries~\citep{fadlon2025imageni2r} discards signal exactly where missingness is highest (App.~\ref{app:masking-tradeoff}). A stochastic imputer restores diversity, but its conditional distribution is fixed before training; as the generator improves, the imputer falls out of sync. Proposition~\ref{prop:frozen-bias} below (mechanism~(iv) in App.~\ref{app:why-close-full}) makes this precise. To state it, let $\xts^{\mathrm{obs}}$ and $\xts^{\mathrm{miss}}$ be the observed and missing parts of a series. For an imputer $q$ over the missing part, the EM objective is $\mathcal{F}(q,\theta) = \mathbb{E}_q[\log p_\theta(\xts^{\mathrm{obs}},\xts^{\mathrm{miss}})] + H(q)$, and it satisfies $\log p_\theta(\xts^{\mathrm{obs}}) = \mathcal{F}(q,\theta) + \mathrm{KL}(q \,\|\, p_\theta(\cdot\mid\xts^{\mathrm{obs}}))$ (App.~\ref{app:em-convergence}). An open-loop pipeline with a frozen imputer $q_0$ trains the generator by maximizing $\mathcal{F}(q_0,\cdot)$.
\begin{proposition}[Irreducible bias of frozen imputation, proof in App.~\ref{app:why-close-full}]\upshape
\label{prop:frozen-bias}
Under standard regularity, let $\theta^\star$ maximize $\log p_\theta(\xts^{\mathrm{obs}})$ and $\theta^\star_{q_0}$ maximize $\mathcal{F}(q_0,\theta)$. Then
\begin{equation*}
\nabla_\theta \log p_\theta(\xts^{\mathrm{obs}}) \big|_{\theta^\star_{q_0}} = \nabla_\theta \mathrm{KL}\big(q_0 \,\|\, p_\theta(\cdot \mid \xts^{\mathrm{obs}})\big) \big|_{\theta^\star_{q_0}},
\end{equation*}
which vanishes under the self-consistency condition $q_0 = p_{\theta^\star_{q_0}}(\cdot \mid \xts^{\mathrm{obs}})$ and is generically non-zero otherwise, and
$0 \le \log p_{\theta^\star}(\xts^{\mathrm{obs}}) - \log p_{\theta^\star_{q_0}}(\xts^{\mathrm{obs}}) \le \mathrm{KL}(q_0 \,\|\, p_{\theta^\star}(\cdot \mid \xts^{\mathrm{obs}})) - \mathrm{KL}(q_0 \,\|\, p_{\theta^\star_{q_0}}(\cdot \mid \xts^{\mathrm{obs}}))$.
\end{proposition}
\noindent In words, training is unbiased \emph{only} when the imputer samples missing values from the same distribution the trained generator itself would: a self-referential requirement that no pre-trained imputer can satisfy, because meeting it would mean knowing the generator in advance. Closing the loop is the unique mechanism that achieves this at every iteration. Iteration is therefore the binding constraint on closed-loop generation under missingness, regardless of the imputer's design.

\vspace{-1mm}
\subsection{Co-evolving MCEM}
\label{subsec:closed-loop}
\vspace{-1mm}

We formalize the closed loop as a latent-variable problem. For each irregular series $\yts^{(i)}$, the complete underlying series $\xts^{(i)}$ is unobserved, and the data likelihood under our generator is an integral over those completions; maximizing it directly is intractable. We therefore use Monte Carlo EM~\citep{dempster1977em,wei1990mcem,neal1998em}, alternating two steps. The \textbf{E-step} samples a complete lifted series from the posterior under the current generator (at iteration $k=0$, we instead use an STL-based initialization; see App.~\ref{app:init}, which also compares other initializations),
\begin{equation}
\label{eq:estep}
\ximg^{(i,k)} \sim p_{\theta_{k-1}}\!\left(\ximg \mid \yts^{(i)}, \Ats^{(i)}\right) \ ,
\end{equation}
and the \textbf{M-step} updates the generator on these completions using the standard EDM~\citep{karras2022edm, naiman2024imagentime} denoising objective with noise-level weighting $\lambda(\sigma)$,
\begin{equation}
\label{eq:mstep}
\theta_k
=
\arg\min_\theta
\mathbb{E}_{i,\sigma,\epsilon}
\left[
\lambda(\sigma)
\left\|
d_\theta(\ximg^{(i,k)}+\sigma\epsilon,\sigma)
-
\ximg^{(i,k)}
\right\|_2^2
\right] \ ,
\qquad
\epsilon \sim \mathcal{N}(0,I) \ .
\end{equation}

The practical benefit of writing the loop this way is concrete: the M-step is identical to clean-data ImagenTime/EDM training: no masked denoising loss, no auxiliary imputation head, and no separate conditional generator. The only difference from standard training is that the completed training set is refreshed between M-steps using the current generator. Substituting a fixed imputer for the E-step reintroduces the frozen-imputer bias (Proposition~\ref{prop:frozen-bias}), so iteration is what resolves it.

\paragraph{What carries over from classical EM.} Classical EM guarantees that the data likelihood never decreases~\citep{dempster1977em,neal1998em}, but only for the idealized loop. We use it as motivation, not as a guarantee for our implementation. Our algorithm meets one of the two assumptions behind it. The M-step only needs to leave the EM objective no worse~\citep{neal1998em}, and our short, warm-started M-step does so at every measured iteration (App.~\ref{app:em-diagnostics}). The E-step, however, is inexact, with one approximate posterior sample per series. Proposition~\ref{prop:frozen-bias} concerns the open-loop baselines and needs no exact E-step.

\vspace{-1mm}
\subsection{Posterior Sampling in Two Spaces}
\label{subsec:lifted-posterior}
\vspace{-1mm}

With the M-step reduced to standard generator training, the remaining challenge is the E-step: at iteration $k$, we must sample complete lifted series $\ximg^{(i,k)}$ via Eq.~\eqref{eq:estep}, i.e., draw from the posterior of the current generator conditioned on the observations. This is a posterior-sampling problem: starting from the prior $p_{\theta_k}(\ximg)$, we seek samples that both follow the learned distribution and are consistent with the observed data. The key difficulty is that the prior and observations live in different spaces. The generator defines a prior over lifted variables $\ximg \sim p_{\theta_k}(\ximg)$, while the observations $\yts$ constrain only the decoded time series $\imgts(\ximg)$. The correct observation model is therefore
\begin{equation}
\label{eq:composed-operator}
G_i \;\triangleq\; \Ats^{(i)} \circ \imgts \ ,
\qquad
\yts^{(i)} = G_i\, \ximg^{(i,k)} \ ,
\end{equation}
which maps an image to a time series. This induces a \emph{dual-space} setting: the diffusion prior is defined in image space, while observations constrain a different domain linked through a decoding map.

Existing diffusion posterior samplers are derived under a single-space assumption, where the observation operator acts directly on the model variable~\citep{chung2023dps, song2023pigdm, zhu2023diffpir, boys2023tmpd, rozet2024diem}. Applying them in the dual-space setting corresponds to conditioning under a misspecified model, replacing the true operator $G_i$ with an approximation that acts in image space. This mismatch is structural and leads to systematic errors. Concretely, three failure modes arise. First, updates are applied with respect to the wrong operator: samples are corrected using an image-space mask rather than the true forward model $G_i$, leading to incorrect update directions. Second, posterior uncertainty is misrepresented: ambiguous regions admit multiple valid completions, but collapsing the lifted representation or treating its coordinates independently averages over modes instead of sampling them. Third, the geometry of the lifted representation is ignored: valid samples lie in $\mathrm{Range}(\tsimg)$, and unconstrained updates can drift outside this subspace, producing images that do not correspond to any valid time series. These issues are not artifacts of a particular sampler but consequences of the dual-space mismatch. We formalize this in App.~\ref{app:mmps-fails}. The next subsection introduces \pslr{}, a minimal recipe that resolves these failures.

\vspace{-1mm}
\subsection{\pslr{}: Posterior Sampling for Lifted Representations}
\label{subsec:pslr}
\vspace{-1mm}

The analysis above implies that posterior sampling in the dual-space setting is well-defined only if three conditions are satisfied: conditioning must be performed through the true observation operator, posterior uncertainty must be preserved, and samples must lie in the space of valid lifted representations. \pslr{} enforces these conditions in a minimal, base-sampler-agnostic way. Importantly, \pslr{} does not modify the underlying posterior-sampling family or the M-step. Instead, it specifies how conditioning must be performed so that the resulting samples correspond to the correct posterior under the dual-space model.

\vspace{-3mm}
\paragraph{C1: operator-consistent conditioning.}
All observation constraints are enforced using the composed operator \( G_i = \Ats^{(i)} \circ \imgts \) from Eq.~\eqref{eq:composed-operator}, rather than an image-space approximation \(\Aimg\). This ensures that updates are computed with respect to the true forward model, correcting the wrong-operator effect described in \cref{subsec:lifted-posterior}.

\vspace{-3mm}
\paragraph{C2: uncertainty preservation.}
Posterior sampling must represent uncertainty over missing values. This requires retaining the full covariance implied by Tweedie's identity (\cref{app:sampler-design-space}), rather than collapsing to a point estimate or a heuristic diagonal, and, in our noiseless setting, regularising the resulting linear system with an adaptive observation-noise scale $\sigma_y = c\,\sigma_t$ so that the posterior remains non-degenerate at every diffusion step (App.~\ref{app:M2}), with $c = 0.10$ throughout (App.~\ref{app:sensitivity}). This preserves multimodality in ambiguous regions instead of averaging over valid completions.

\vspace{-3mm}
\paragraph{C3: manifold consistency.}
Samples must lie in $\mathrm{Range}(\tsimg)$. Our lift is overcomplete: each time step is stored in several pixels (App.~\ref{app:lift-delay}), and only images whose copies agree correspond to a time series. The base sampler updates pixels freely and cannot keep the copies in agreement. We enforce this by projection,
\[
\ximg \leftarrow \Pi(\ximg) = \tsimg(\imgts(\ximg)) \ ,
\]
which removes off-manifold components before passing samples to the M-step. Since decoding averages the copies of each time step, $\Pi$ leaves valid images unchanged and only corrects inconsistent ones. Between the decode and the re-lift, we also write the observed values back into the series (\cref{alg:em-compact}). We do this in series space because overwriting single pixels would break the agreement between copies again. This prevents the accumulation of invalid structure across EM iterations.

\medskip
\noindent
Together, C1--C3 ensure that posterior sampling is performed under the correct dual-space model. The recipe is base-sampler-agnostic: any sampler that exposes posterior uncertainty can be adapted by enforcing these conditions. In practice, we instantiate \pslr{} with MMPS~\citep{rozet2024diem}, which already preserves posterior covariance, yielding our primary sampler $\lmmps \triangleq \pslr{}(\mathrm{MMPS})$. Applying the same construction to TMPD~\citep{boys2023tmpd} yields $\ltmpd$, used to verify transfer across samplers (\cref{app:ltmpd}). Implementation details are given in \cref{alg:estep} and App.~\ref{app:mmps-fails}. \Cref{tab:failure-modes} in App.~\ref{app:mmps-fails} summarizes the three failure modes, the condition that fixes each one, and the score without that condition.

\begin{figure*}[t]
\centering
\includegraphics[width=\textwidth]{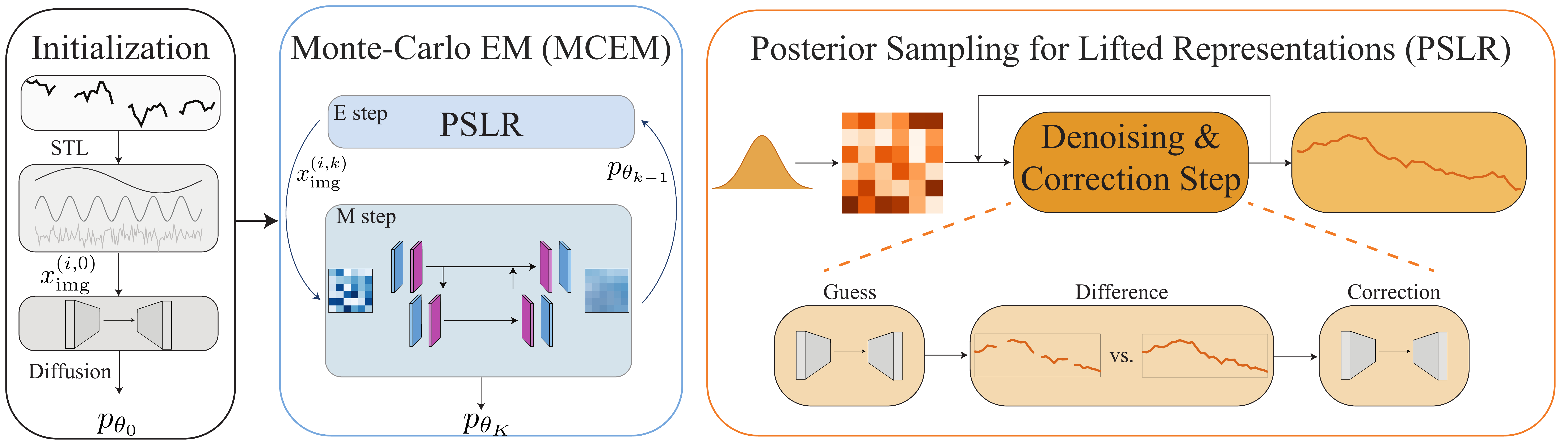}
\vspace{-5mm}
\caption{Overview of our framework. We initialize a generator from a simple decomposition of irregular time series, then refine it via a co-evolving Monte Carlo EM (MCEM) loop that alternates between posterior sampling (E-step) and generator updates (M-step). Posterior Sampling for Lifted Representations (PSLR) enables conditioning in the lifted space by iteratively denoising and correcting samples to match the observations.}
\label{fig:method}
\vspace{-2mm}
\end{figure*}

\vspace{-1mm}
\subsection{Full algorithm}
\label{subsec:algorithm}
\vspace{-1mm}

\begin{algorithm}[t]
\small
\caption{Co-evolving EM with a lifted posterior-sampling E-step}
\label{alg:em-compact}
\begin{algorithmic}[1]
\Require irregular data $\{(\yts^{(i)},\Ats^{(i)})\}_{i=1}^N$; lift $\tsimg$ and inverse $\imgts$; initializer $\mathcal{I}$; EM iterations $K$
\State \textbf{Init:} $\ximg^{(i,0)} \gets \tsimg(\mathcal{I}(\yts^{(i)},\Ats^{(i)}))$ for all $i$
\State Train initial prior $p_{\theta_0}(\ximg)$ on $\{\ximg^{(i,0)}\}_{i=1}^N$ using the EDM objective
\For{$k=1,\ldots,K$}
    \State \textbf{E-step:} with $\theta_{k-1}$ frozen, sample
    \[
    \ximg^{(i,k)} \sim p_{\theta_{k-1}}(\ximg \mid \yts^{(i)},\Ats^{(i)})
    \]
    using \pslr{} with $G_i=\Ats^{(i)}\circ\imgts$
    \State Project: decode $\tilde{x} \gets \imgts(\ximg^{(i,k)})$, set observed entries of $\tilde{x}$ to $\yts^{(i)}$, re-lift $\ximg^{(i,k)} \gets \tsimg(\tilde{x})$
    \State \textbf{M-step:} update $\theta_k$ by the EDM objective in Eq.~\eqref{eq:mstep} on $\{\ximg^{(i,k)}\}_{i=1}^N$
\EndFor
\State \textbf{Inference:} sample $\ximg\sim p_{\theta_K}(\ximg)$ with standard EDM and return $\imgts(\ximg)$
\end{algorithmic}
\end{algorithm}

\Cref{alg:em-compact} and~\cref{fig:method} summarize the full procedure end-to-end. The loop has a simple interpretation: the E-step asks “given the current generator, what complete series are plausible under the observations?”, while the M-step asks “given these completions, what generator best explains them?”. This feedback is precisely what is missing from open-loop pipelines.

The overall framework is modular. The \textbf{lift} $\tsimg$ can be any linear map with a (pseudo-)inverse (delay embedding by default, STFT for validation); the \textbf{operator} $\Ats$ can represent arbitrary observation processes (random masks, block gaps, fractional resampling); the \textbf{E-step sampler} can be any posterior sampler satisfying C1–C3; and the \textbf{backbone} can be any differentiable denoiser (EDM in our implementation, but also flow-matching or consistency models). This modularity allows the same closed-loop procedure to extend across architectures and observation models.

App.~\ref{app:mmps-fails} derives the three structural failure modes (M1--M3) that motivate C1--C3, and the ablation in \cref{sec:ablation} validates each one empirically. We now turn to experiments validating each component across ten datasets and four stress tests.

\vspace{-2mm}
\section{Results}
\label{sec:results}
\vspace{-1mm}

\vspace{-1mm} \subsection{Quantitative Evaluation} \label{sec:quant-results} \vspace{-1mm} All prior work on irregular time-series generation~\citep{yoon2019timegan,jeon2022gtgan,fadlon2025imageni2r} uses the same corruption pattern: a fixed fraction of timesteps is dropped at random time steps in the sequence, every feature at a dropped timestep goes missing together, and observations are taken at evenly spaced times (\cref{sec:quant-standard}). Real-world time series are typically more complex: sensors fail for continuous periods, channels fail asynchronously, and sampling clocks are rarely uniform. We close this gap by introducing three new benchmarks alongside the standard one: \textbf{block-structured} missingness, \textbf{per-cell} asynchronous missingness, and \textbf{continuous-time} irregular sampling. For each new benchmark, we evaluate on at least four datasets from~\citep{fadlon2025imageni2r} (ETTh1, ETTh2, ETTm1, ETTm2, Weather, Electricity, Energy, Sine, Stock, Mujoco) and the four standard metrics: discriminative score, predictive score, Context-FID, and feature correlation (plus downstream classification, App.~\ref{app:downstream}). In irregular time-series generation, models are trained exclusively on corrupted observations, while evaluation is performed by comparing unconditionally generated samples to the fully observed data distribution using the metrics above. We compare our method against ImagenI2R~\citep{fadlon2025imageni2r}, KoVAE~\citep{naiman2024kovae}, GT-GAN~\citep{jeon2022gtgan}, and TimeGAN-$\Delta t$~\citep{yoon2019timegan,che2018recurrent}.

Across all four settings, our EM framework achieves SOTA results (see \cref{tab:avg-benchmarks}), while prior methods, built around the standard closed-loop pattern, break down on the three new benchmarks. On average, we achieve relative improvements of 59\%, 15\%, 73\%, and 71\% in discriminative score, predictive score, Context-FID, and feature correlation, respectively. The new regimes expose complementary failure modes of open-loop imputation. In Block, long contiguous gaps substantially widen the conditional distribution, so point imputers collapse multiple plausible trajectories into oversmoothed averages. In Cell, predictive information is distributed across simultaneously observed channels, but independent or weakly coupled imputers cannot fully exploit this cross-feature structure. Because these completions are frozen before generator training, the resulting bias becomes embedded in the learned generative distribution. The use of an EM framework may appear to be a drawback due to potentially long training times, but in practice convergence is fast and this concern does not materialize: averaged across all benchmarks, our method reduces time-to-best discriminative score by $58\%$ compared to KoVAE, while remaining within a $2.2\times$ factor of ImagenI2R. On average $\sim\!5$ EM iterations are required to reach optimal performance, and each M-step uses only $40$ training epochs. Crucially, each M-step is initialized from the previous iteration's weights rather than retrained from scratch, making training incremental and yielding substantial wall-clock savings (for more info refer to App.~\ref{app:convergence}). A full description of the datasets, metrics and previous methods is in App.~\ref{app:setup-full}. Full per-dataset and per-configuration results are provided in App.~\ref{app:additional-exp}.
\begin{table}[t]
    \centering
    \small
    \setlength{\tabcolsep}{3.5pt}
    \setlength{\aboverulesep}{1pt}
    \setlength{\belowrulesep}{1.5pt}
    \caption{Averaged results across datasets for each benchmark.}
    \label{tab:avg-benchmarks}
    \scalebox{0.85}{
    \begin{tabular}{ll|ccccc}
        \toprule
        \textbf{Metric} & \textbf{Setting}
        & \textbf{TimeGAN-$\Delta t$} & \textbf{GT-GAN} & \textbf{KoVAE} & \textbf{ImagenI2R} & \textbf{Ours} \\
        \midrule

        \multirow{4}{*}{\textbf{Disc.} $\downarrow$}
        & Timestep (Standard)   & 0.496 & 0.387 & 0.228 & 0.067 & \cellcolor{blue!10}$\boldsymbol{0.055}$ \\
        & Blocking   & 0.499 & 0.432 & 0.321 & 0.366 & \cellcolor{blue!10}$\boldsymbol{0.097}$ \\
        & Cell       & 0.494 & 0.447 & 0.341 & 0.304 & \cellcolor{blue!10}$\boldsymbol{0.116}$ \\
        & Continuous & 0.498 & 0.427 & 0.326 & 0.274 & \cellcolor{blue!10}$\boldsymbol{0.043}$ \\

        \midrule
        \multirow{4}{*}{\textbf{Pred.} $\downarrow$}
        & Timestep (Standard)   & 0.317 & 0.168 & 0.150 & 0.132 & \cellcolor{blue!10}$\boldsymbol{0.109}$ \\
        & Blocking   & 0.377 & 0.139 & 0.131 & 0.155 & \cellcolor{blue!10}$\boldsymbol{0.110}$ \\
        & Cell       & 0.363 & 0.193 & 0.130 & 0.138 & \cellcolor{blue!10}$\boldsymbol{0.108}$ \\
        & Continuous & 0.339 & 0.186 & 0.114 & 0.115 & \cellcolor{blue!10}$\boldsymbol{0.104}$ \\

        \midrule
        \multirow{4}{*}{\textbf{FID} $\downarrow$}
        & Timestep (Standard)   & 4.355 & 5.172 & 6.587 & 0.373 & \cellcolor{blue!10}$\boldsymbol{0.296}$ \\
        & Blocking   & 6.050 & 8.188 & 3.971 & 4.893 & \cellcolor{blue!10}$\boldsymbol{0.657}$ \\
        & Cell       & 6.442 & 7.598 & 3.938 & 3.595 & \cellcolor{blue!10}$\boldsymbol{0.417}$ \\
        & Continuous & 6.062 & 4.075 & 3.618 & 3.057 & \cellcolor{blue!10}$\boldsymbol{0.030}$ \\

        \midrule
        \multirow{4}{*}{\textbf{Corr.} $\downarrow$}
        & Timestep (Standard)   & 2.664 & 2.911 & 1.332 & 0.330 & \cellcolor{blue!10}$\boldsymbol{0.250}$ \\
        & Blocking   & 3.460 & 3.681 & 2.394 & 1.879 & \cellcolor{blue!10}$\boldsymbol{0.227}$ \\
        & Cell       & 4.399 & 4.521 & 2.915 & 2.182 & \cellcolor{blue!10}$\boldsymbol{0.362}$ \\
        & Continuous & 4.062 & 4.699 & 3.204 & 1.906 & \cellcolor{blue!10}$\boldsymbol{0.243}$ \\

        \bottomrule
    \end{tabular}
    }
\end{table}

\paragraph{Standard benchmark: Fix missing rates}
\label{sec:quant-standard}

Following the setting of~\citep{yoon2019timegan,jeon2022gtgan}, we randomly remove $30\%$, $50\%$, and $70\%$ of the observations from each training sample. In this setting, missingness is applied at the timestep level: if a timestamp is missing, all features at that timestamp are considered missing. We evaluate this benchmark on all 10 datasets mentioned above, across three sequence lengths $\{24, 96, 768\}$ and three missing rates.

\vspace{-2mm}
\paragraph{Block-structured missingness.}
\label{sec:quant-block}

Sensor failures often produce contiguous blocks of missing values driven by physical mechanisms, e.g., a wind farm shutting down during a storm or a weather station going offline during a cold snap, a setting that challenges interpolation-based imputers. We evaluate three blocking patterns: block-start, block-random, and block-end at missing rates of $30\%$, $50\%$, and $70\%$ on the datasets mentioned above except Electricity and Mujoco ($L{=}24$). For example, in the block-start setting with a $50\%$ missing rate, we remove the first half of the sequence, creating a contiguous block of missing observations at the beginning. \looseness=-1

\vspace{-2mm}
\paragraph{Per-cell (asynchronous-channel) missingness}
\label{sec:quant-cell}

The standard benchmark removes entire timesteps (row-level missingness), assuming that all features at a given timestamp are either fully observed or fully missing. In practice, however, sensors monitoring different channels fail independently, so one feature may be observed while another at the same timestep is missing. To model this, we drop each $(t, f)$ entry independently with probability $\rho$, resulting in an asynchronous, per-feature missingness pattern. For example, at a given timestep $t$, the temperature sensor may be missing while the humidity and pressure sensors remain observed. We evaluate this setting on ETTh1, ETTm1, Stock, Energy, and Weather at missing rates of $30\%$, $50\%$, and $70\%$.

\vspace{-2mm}
\paragraph{Continuous-time irregular sampling}
\label{sec:quant-continuous}

The previous benchmarks assume observations are taken on a regular time grid (e.g., $t=0,1,2,\dots$). In many real-world settings, however, measurements are collected at irregular, continuous times (e.g., $t=0.3, 1.7, 3.25$). To simulate this, we start from a clean sequence of length $T$ and sample $N_{\mathrm{obs}}=\mathrm{round}(\rho T)$ real-valued timestamps uniformly in $[0, T{-}1]$, sorted, independently per sample, where $\rho$ is the \emph{observation density} ($\rho{=}0.5$ keeps half the grid points). The value at each sampled time is obtained by linear interpolation between the two nearest grid points. For example, if $t=3.25$, the observation is $0.75 \cdot x_3 + 0.25 \cdot x_4$. As a result, the model observes only a set of $(t_k, y_k)$ pairs at irregular times, rather than values on a fixed grid. During training, the model is exposed only to these irregular observations, while at evaluation time it generates full sequences on the original grid. We evaluate on ETTh1, Energy, Weather, and Stock for $\rho \in \{0.3, 0.5\}$. \looseness=-1

\vspace{-1mm}
\subsection{Qualitative evaluation}
\label{sec:qualitative}
\vspace{-1mm}

To qualitatively assess the fidelity of generated sequences, we follow \citep{yoon2019timegan} and use two visualization techniques: t-SNE \citep{van2008visualizing} to project real and synthetic data into a 2D space, and kernel density estimation to visualize the underlying PDFs. \cref{fig:70_tsne_density} (App.~\ref{sec:qual_eval_full}) shows results for $70\%$ missingness across Weather, Energy, and Stocks, where the top row presents 2D distributions for real data (deep blue), our method (vermillion, dashed), and ImagenI2R (teal, dashed), and the bottom row shows the corresponding PDFs. Our model closely matches the real data, with near-perfect overlap in t-SNE and well-aligned densities, while ImagenI2R appears more concentrated and slightly biased. Additional results for $30\%$ and $50\%$ missingness, along with Wasserstein distances, are provided in App.~\ref{sec:qual_eval_full}.

\subsection{Ablation Study}
\label{sec:ablation}

\paragraph{Stochastic imputation and extended training are not enough.}
A natural objection to our critique of point imputers is that their failure may stem from deterministic averaging rather than the open-loop pipeline itself. We test this by evaluating the \textbf{Point imputer (NCDE) + ImagenTime} baseline, which represents the traditional open-loop approach where a Neural CDE generates point completions once before training. Next, we replace the deterministic imputer (NCDE) with a stochastic one (CSDI). While stochasticity restores diversity and halves the error (0.218 $\rightarrow$ 0.095), it remains a suboptimal ceiling because the completions are frozen before training begins. To isolate the effect of iteration from raw compute, we evaluate \textbf{Ours (1 iter, extended training)}, which draws a single set of posterior samples and trains for the full epoch budget without refreshing. This baseline performs similarly to CSDI (0.115), proving that a single sampling step is fundamentally insufficient. The massive performance leap achieved by the full \textbf{Ours (EM + \pslr{})} framework (0.027) confirms that continuously refining imputations with generator feedback is the true binding constraint for generative fidelity. \looseness=-1

\paragraph{Each \pslr{} component matters in dual-space.}
We further ablate the structural components of our lifted posterior-sampling recipe. Simply closing the loop using a naive, single-space posterior sampler (\textbf{Lifted-mask MMPS-EM}) yields poor results (0.087), proving that ignoring cross-space constraints prevents proper convergence. Removing \textbf{C1} (operator-consistent conditioning) degrades accuracy (0.038), demonstrating that evaluating constraints via image-space approximations leads to incorrect update directions. Removing \textbf{C2} (uncertainty preservation) hurts performance (0.047), consistent with the failure mode analyzed in App.~\ref{app:M2}. Removing \textbf{C3} (manifold consistency) causes off-manifold noise to accumulate across EM iterations, further degrading performance (0.056). The \textbf{Full \pslr{}-EM} successfully integrates these corrections to bridge the time-series and image domains, effectively closing the gap to the \textbf{Clean-data oracle} (0.027 vs. 0.007).

\begin{table}[t]
\centering
\setlength{\tabcolsep}{3.5pt}
\setlength{\aboverulesep}{1pt}
\setlength{\belowrulesep}{1.5pt}
\caption{Ablation study reporting average Disc. Scores across Energy, Weather, and Stock datasets (30\%, 50\%, 70\% drop rates). The first block isolates the impact of the iterative feedback loop, the second evaluates structural components, and the final block shows the optimized performance against the clean-data oracle. Lower is better.}
\label{tab:ablation}
\resizebox{\linewidth}{!}{
\begin{tabular}{l|c|c|c|ccc|c}
\toprule
\textbf{Method}
& \textbf{Closed-loop}
& \textbf{Imputer}
& \textbf{PSLR}
& \textbf{Energy}
& \textbf{Weather}
& \textbf{Stock}
& \textbf{Avg.} \\
\midrule
Point imputer (NCDE) + ImagenTime & \xmark & Deterministic & \xmark & 0.225 & 0.281 & 0.149 & 0.218 \\
CSDI + ImagenTime & \xmark & Stochastic & \xmark & 0.120 & 0.095 & 0.070 & 0.095 \\
Ours (1 iter, extended training) & \xmark & Posterior & \checkmark & 0.140 & 0.120 & 0.085 & 0.115 \\
\midrule
Lifted-mask MMPS-EM & \checkmark & Posterior & \xmark & 0.095 & 0.098 & 0.096 & 0.087 \\
w/o C1 & \checkmark & Posterior & partial & 0.065 & 0.035 & 0.014 & 0.038 \\
w/o C2 & \checkmark & Posterior & partial & 0.072 & 0.048 & 0.022 & 0.047 \\
w/o C3 & \checkmark & Posterior & partial & 0.078 & 0.061 & 0.031 & 0.056 \\
\midrule
\textbf{Ours (Full PSLR-EM)} & \checkmark & Posterior & \checkmark & \cellcolor{blue!10}$\boldsymbol{0.051}$ & \cellcolor{blue!10}$\boldsymbol{0.022}$ & \cellcolor{blue!10}$\boldsymbol{0.007}$ & \cellcolor{blue!10}$\boldsymbol{0.027}$ \\
Clean-data oracle & - & - & - & 0.009 & 0.003 & 0.007 & 0.006 \\
\bottomrule
\end{tabular}
}
\end{table}

\vspace{-2mm}
\section{Conclusions}
\label{sec:conclusions}
\vspace{-1mm}

Regular time-series generation from irregular inputs is limited less by model class than by training dynamics: open-loop pipelines fit a generator to a stale surrogate produced by a frozen imputer. We close the loop with MCEM in lifted space and introduce \pslr{}, an E-step recipe enforcing operator-consistent conditioning, uncertainty preservation, and manifold consistency (C1--C3). The resulting plug-and-play method recovers near-oracle performance across four corruption regimes, suggesting E-step geometry, not generator capacity, is the binding design choice for generative learning under missingness. Three limitations remain. First, the observed data do not always determine the true distribution. They do when each value is dropped independently, but when whole timesteps are dropped they only determine short windows of the series (Remark~\ref{rem:identifiability}). In that case, and under long gaps or value-dependent missingness (App.~\ref{app:mnar}), our model can match the observed data and still differ from the truth. Our robustness results also hold only for the settings we tested. Second, EM only converges to a local optimum, which means the final solution depends on initialization. Finally, although our framework is faster than NCDE- and CSDI-based pipelines, it is not in the cheapest compute tier: wall-clock is roughly $2.2\times$ that of the fastest single-pass diffusion baseline, ImagenI2R (\cref{tab:wallclock-comparison}). Future work targets these directly: amortising posterior sampling across iterations to narrow the runtime gap, extending \pslr{} to nonlinear lifts, and pairing the EM fixed point with a maximum-entropy regulariser to tighten identifiability. \looseness=-1

\begin{ack}
This research was partially supported by the Lynn and William Frankel Center of The Stein Faculty of Computer and Information Science, Ben-Gurion University of the Negev, and by an ISF grant 1299/25. The authors declare no competing interests.
\end{ack}

\clearpage
\bibliographystyle{abbrvnat}
\bibliography{refs_camera_ready}

\clearpage
\appendix

\section{Time Series-to-Image Lift}
\label{app:lift}

In this section we provide three representative time series-to-image transformations. Delay embedding is the default used throughout the paper, STFT is used only as the validation lift of App.~\ref{app:stft}, and vanilla folding is included as the simplest pedagogical baseline (not used in any of our experiments). For each, we give the forward map $\tsimg$ and its (pseudo-)inverse $\imgts$. Throughout, $\xts \in \mathbb{R}^{\Nts \times F}$ denotes a multivariate series of length $\Nts$ with $F$ features, and $\ximg = \tsimg(\xts)$ its image lift.

\subsection{Delay embedding}
\label{app:lift-delay}

Delay embedding~\citep{takens1981embedding} converts a univariate channel $x_{1:\Nts}$ into an image whose columns are time-shifted slices of the input. Two hyperparameters control the geometry: the embedding dimension $n$ (number of rows) and the skip $m$ (column stride). The forward map produces the matrix
\[
X \;=\; \begin{bmatrix}
x_1     & x_{m+1}    & \cdots & x_{\Nts - n} \\
\vdots  & \vdots     & \ddots & \vdots \\
x_n     & x_{n+m+1}  & \cdots & x_{\Nts}
\end{bmatrix}
\;\in\; \mathbb{R}^{n \times q}, \qquad q = \lceil (\Nts - n)/m \rceil,
\]
which is then zero-padded to $n \times n$ to match the convolutional backbone. For multivariate input with $F$ features, per-channel matrices are stacked along the channel axis, giving $\ximg \in \mathbb{R}^{F \times n \times n}$. The lift scales gracefully to long sequences: choosing $m = n = 256$ encodes series of length up to $65\,$k in a single $256 \times 256$ image: the scaling that motivates this lift on the long-sequence benchmarks of~\citep{naiman2024imagentime}.

\paragraph{Inverse:}
When $m < n$, delay embedding is \emph{overcomplete}: a given time index $i$ appears in several image pixels. We define $\imgts(\ximg)[i]$ as the \emph{average} of all image pixels that delay-embed to index $i$. This averaging inverse is the Moore--Penrose pseudo-inverse of $\tsimg$ and yields perfect round-trip reconstruction $\imgts(\tsimg(\xts)) = \xts$.

\paragraph{Why the pseudo-inverse matters for \pslr{}.}
The Moore--Penrose property is what allows the three E-step corrections of \cref{subsec:pslr} to be applied uniformly across lifts. First, the composition $\Pi = \tsimg \circ \imgts$ is then the orthogonal projector onto $\mathrm{Range}(\tsimg)$, so manifold consistency (C3) is enforced by a single idempotent application of $\Pi$. Second, operator-consistent conditioning (C1) requires the adjoint $G^\top v = \tsimg(\Ats^\top v)$ of the composed operator $G_i = \Ats^{(i)} \circ \imgts$, which under the averaging inverse is the broadcast that writes each time-series entry into all of its image pixels divided by the multiplicity, a sparse per-pixel scale that costs no more than one forward lift.

\subsection{Short-time Fourier transform (STFT)}
\label{app:lift-stft}

The STFT~\citep{griffin1984signal} maps a time series to a time--frequency image by sliding a fixed-length window across it and applying a Fourier transform inside each window. Concretely, given a univariate channel $x_{1:\Nts}$, an analysis window $w[\cdot]$ of length $n_{\mathrm{fft}}$, and a hop length $h$, the forward map computes
\[
\mathrm{STFT}(x)[k, j] \;=\; \sum_{n = 0}^{n_{\mathrm{fft}} - 1} x[jh + n]\, w[n]\, e^{-2\pi i\, k n / n_{\mathrm{fft}}}, \qquad k = 0, \ldots, n_{\mathrm{fft}}/2,\; j = 0, \ldots, W - 1.
\]
Each column $j$ is the local frequency content of the window centred at time $jh$; columns therefore preserve a notion of locality in time, in contrast to vanilla folding. The output is complex-valued; we follow~\citep{naiman2024imagentime} and stack the real and imaginary parts as separate channels, yielding $\ximg \in \mathbb{R}^{2F \times H \times W}$ with $H = n_{\mathrm{fft}}/2 + 1$. Unlike standard audio practice we do \emph{not} compute the magnitude spectrogram: keeping both real and imaginary components avoids the ill-posed phase-reconstruction step at inference. We rescale the resulting image to $[-1, 1]$ before passing it to the diffusion backbone.

\paragraph{Inverse:}
The inverse short-time Fourier transform (ISTFT) reconstructs the time series in closed form provided the analysis window pair satisfies the constant-overlap-add (COLA) condition $\sum_j w[n - jh]^2 = \mathrm{const}$ for all $n$~\citep{griffin1984signal}: under COLA, ISTFT is a linear map and yields perfect reconstruction. We use the ImagenTime STFT configuration ($n_{\mathrm{fft}}{=}8$, $h{=}4$, Hann window) for the lift-agnosticism experiments of App.~\ref{app:stft}; the Hann window with $50\%$ overlap satisfies COLA, so $\mathcal{T}^{-1}_{\mathrm{STFT}}$ is closed-form linear.

\paragraph{Linearity remark.}
All three lifts above (folding, delay embedding, STFT after the real/imag split) are linear maps with closed-form pseudo-inverses (each output pixel is a fixed linear combination of input entries), so the three E-step corrections (C1--C3) of \cref{subsec:pslr} apply verbatim regardless of the lift.

\section{PSLR E-Step and STL Warm-Start}
\label{app:algorithms}

Algorithm~\ref{alg:estep} details the E-step (\pslr{} sample); the outer loop is Algorithm~\ref{alg:em-compact} from the main text.

\begin{algorithm}[h]
\caption{$\lmmps$ E-step (single sample)}
\label{alg:estep}
\begin{algorithmic}[1]
\Require observation $\yts \in \mathbb{R}^{\Nobs \times F}$, mask $\Ats$
\Require denoiser $d_\theta$, EDM noise schedule $\{\sigma_t\}_{t = T, \ldots, 0}$
\Require initial image $x_{\mathrm{init}} \in \mathbb{R}^{\Nimg}$ (from the STL warm start, App.~\ref{app:init}, at $k{=}1$; from the previous E-step output at $k{>}1$)
\Ensure posterior sample $\ximg \in \mathrm{Range}(\tsimg)$ with $G\,\ximg \approx \yts$
\State $x_T \gets x_{\mathrm{init}} + \sigma_T \epsilon$, \quad $\epsilon \sim \mathcal{N}(0, I)$
\For{$t = T, T-1, \ldots, 1$}
  \State $\hat{x}_0 \gets d_\theta(x_t, \sigma_t)$ \Comment{Tweedie mean}
  \State $\sigma_y \gets c \cdot \sigma_t$ \Comment{C2: uncertainty preservation, adaptive $\sigma_y$ (\cref{subsec:pslr})}
  \State $r \gets \yts - \Ats \cdot \imgts(\hat{x}_0)$ \Comment{C1: operator-consistent residual via true $G$}
  \State $v \gets \mathrm{CG}_{\mathrm{obs}}(r; \sigma_y, \sigma_t, d_\theta, x_t, n_{\mathrm{CG}})$ \Comment{runs in $\mathbb{R}^{\Nobs \times F}$}
  \State $\delta \gets \mathrm{VJP}(d_\theta, x_t, \tsimg(\Ats^\top v))$ \Comment{posterior score: $J^\top G^\top v$}
  \State $x_{t-1} \gets \mathrm{EDMstep}(x_t, \hat{x}_0 + \sigma_t^2 \delta, \sigma_t, \sigma_{t-1})$
  \If{$t \mod p_{\mathrm{proj}} = 0$} \Comment{C3: manifold consistency, projection by $\Pi$}
    \State $\tilde{x}_{\mathrm{ts}} \gets \imgts(x_{t-1})$
    \State $\tilde{x}_{\mathrm{ts}}[\mathrm{observed}] \gets \yts[\mathrm{observed}]$
    \State $x_{t-1} \gets \tsimg(\tilde{x}_{\mathrm{ts}})$
  \EndIf
\EndFor
\State \Return $x_0$
\\
\Function{$\mathrm{CG}_{\mathrm{obs}}$}{$r; \sigma_y, \sigma_t, d_\theta, x_t, n_{\mathrm{CG}}$}
  \State Define $A_{\mathrm{mvp}}(v_{\mathrm{ts}})$:
  \State \quad $v_{\mathrm{img}} \gets \tsimg(\Ats^\top v_{\mathrm{ts}})$ \Comment{$G^\top v$}
  \State \quad $J v_{\mathrm{img}} \gets \mathrm{VJP}(d_\theta, x_t, v_{\mathrm{img}})$
  \State \quad $J v_{\mathrm{ts}} \gets \Ats \cdot \imgts(J v_{\mathrm{img}})$ \Comment{$G(\cdot)$}
  \State \quad \Return $\sigma_y^2\, v_{\mathrm{ts}} + \sigma_t^2\, J v_{\mathrm{ts}}$
  \State \Return $\mathrm{CG}(A_{\mathrm{mvp}}, r, \text{max iters} = n_{\mathrm{CG}})$
\EndFunction
\end{algorithmic}
\end{algorithm}

\paragraph{Warm-start construction (STL).}
\label{app:init}
Iteration $0$ has no generator to sample from, the first M-step must train on an externally supplied completion. Without structure, that completion drives the first M-step toward corruption artefacts and the first E-step off the data manifold (App.~\ref{app:mmps-fails}). We seed $\theta_0$ with a closed-form, parameter-free completion that injects the canonical trend / seasonal / residual decomposition via a single STL pass~\citep{cleveland1990stl}: at missing positions we keep the trend and seasonal components and resample residuals from the observed-residual moments. The warm start's role is structural; the EM loop closes the residual gap to the data distribution, and convergence is empirically insensitive to the specific structurally-correct initialiser.

\paragraph{Impact of the warm start.} To measure how much the warm start matters, we ran the full pipeline at $50\%$ missing on Weather and Energy with four initializations of increasing quality: random noise, linear interpolation, Kalman smoothing, and STL (our default). Everything else is unchanged. \Cref{tab:warm-start} reports the best discriminative score and the number of EM iterations needed to reach it. The two structured starts, STL and Kalman, reach nearly the same score at the same speed on both datasets. The random start needs about twice as many iterations and has not fully converged on Weather within the budget, and the linear start converges well on Weather but slowly on Energy. In these tests, a better start mainly saves EM iterations, while the final score is set by the EM loop itself.

\begin{table}[ht]
\centering
\caption{Warm-start sensitivity at $50\%$ missing. Best discriminative score ($\downarrow$) and EM iterations needed to reach it. STL is the default used in all other experiments.}
\label{tab:warm-start}
\small
\begin{tabular}{lcccc}
\toprule
 & \multicolumn{2}{c}{Weather} & \multicolumn{2}{c}{Energy} \\
\cmidrule(lr){2-3} \cmidrule(lr){4-5}
Initialization & Disc. & Iters. & Disc. & Iters. \\
\midrule
Random noise & 0.057 & $>7$ & 0.045 & 10 \\
Linear interpolation & 0.026 & 4 & 0.157 & $>17$ \\
Kalman smoothing & 0.021 & 3 & \textbf{0.040} & 5 \\
STL (default) & \textbf{0.020} & 3 & 0.042 & 5 \\
\bottomrule
\end{tabular}
\end{table}

\section{Theoretical Background and Analysis}
\label{app:theory}

This section consolidates the theoretical content of the paper into four subsections. \ref{app:sampler-design-space} reviews the posterior-sampling design space (DPS, $\Pi$GDM, DiffPIR, TMPD, MMPS) for non-time-series readers. \ref{app:em-convergence} reviews classical EM convergence (free energy, Neal--Hinton monotonicity, MCEM). \ref{app:why-close-full} consolidates the four mechanisms that make iteration necessary in the open-loop $\to$ closed-loop transition, including Proposition~\ref{prop:frozen-bias}. \ref{app:mmps-fails} derives the three structural misspecifications (M1--M3) that any single-space Tweedie-moment sampler exhibits in the dual-space setting, and motivates the corresponding \pslr{} corrections (C1--C3).

\subsection{Posterior-sampling design space}
\label{app:sampler-design-space}

All Tweedie-moment posterior samplers start from Bayes' rule on the score:
\[
\nabla_{x_t} \log p(x_t \mid y) \;=\; \nabla_{x_t} \log p(x_t) \;+\; \nabla_{x_t} \log p(y \mid x_t).
\]
The prior score is provided by the denoiser. The challenge is approximating the likelihood score, equivalently the posterior covariance $\mathbb{V}[x \mid x_t]$. Each sampler in the design space makes a different approximation; \cref{tab:samplers-motivation} summarises the choices and the dual-space failure modes derived in App.~\ref{app:mmps-fails}. Throughout, every method assumes an observation model $y = Ax + \mathrm{noise}$ where $A$ operates in the model's native space; in our setting the true model is $\yts = G \, \ximg$ with $G = \Ats \circ \imgts$.

\begin{table}[ht]
\centering
\caption{Tweedie-moment posterior-sampler design space. Approximations of the posterior covariance $\mathbb{V}[x \mid x_t]$ and the structural failure that each sampler exhibits when dropped unchanged into the dual-space lifted setting (App.~\ref{app:mmps-fails}). DPS, $\Pi$GDM, and DiffPIR are structurally incompatible with our \pslr{} recipe (they violate C2 or admit no closed-form proximal step under the composed operator $G$) and are therefore not assigned a discriminative score. Ablation column lists the corresponding row of \cref{tab:ablation} on Energy $50\%$ for the samplers that can be slotted into our E-step.}
\label{tab:samplers-motivation}
\resizebox{\textwidth}{!}{
\begin{tabular}{l|c|c|l|c}
\toprule
\textbf{Sampler} & \textbf{Posterior covariance} & \textbf{Linear solver} & \textbf{Dual-space failure mode} & \textbf{Ablation cell} \\
\midrule
DPS~\citep{chung2023dps}        & point estimate (delta)              & --                       & MAP-like, collapses posterior diversity         & ---             \\
$\Pi$GDM~\citep{song2023pigdm}  & isotropic diagonal $\mathrm{diag}(v_t)$ & --                  & no inter-pixel correlation, drops lift redundancy & ---             \\
DiffPIR~\citep{zhu2023diffpir}  & none (proximal)                     & inner solver collapses    & no closed form once $A$ is composed with $\imgts$ & ---             \\
TMPD~\citep{boys2023tmpd}       & row-sum diagonal $\mathrm{diag}(\mathbf{1}^\top J)$ & per-step closed form & still drops block correlation; converges-then-diverges & 0.112 \\
MMPS~\citep{rozet2024diem}      & full Jacobian $\sigma_t^2 J$        & CG (image-space)          & wrong operator $\Aimg$, not $G$; biased fixed point & $0.095$ \\
\midrule
\textbf{$\lmmps$ (ours)}        & full Jacobian $\sigma_t^2 J$        & \textbf{CG (obs-space)}   & \textbf{none (regime a)}                            & $\mathbf{0.045}$ \\
\bottomrule
\end{tabular}
}
\end{table}

\paragraph{DPS~\citep{chung2023dps}.}
Replaces $p(x \mid x_t)$ with a delta function at the Tweedie mean, a zeroth-order approximation that ignores covariance entirely. Recent analysis~\citep{xu2025rethinking} shows DPS performs MAP estimation rather than posterior sampling, producing low-diversity outputs. In EM, posterior diversity is essential: MAP-like behaviour collapses completions to a single point per observation, preventing the M-step from learning distributional structure. DPS is therefore structurally incompatible with our \pslr{} recipe (it violates C2, full Tweedie covariance) and is not assigned a discriminative score.

\paragraph{$\Pi$GDM~\citep{song2023pigdm}.}
Diagonal covariance: $\mathbb{V}[x \mid x_t] \approx \mathrm{diag}(v_t)$. Delay embedding creates structured off-diagonal correlations in $\mathbb{V}[\ximg \mid x_t]$ (each time point maps to $d$ overlapping image positions); a diagonal approximation treats these copies as independent, underestimating manifold-direction covariance by $\sim d$. $\Pi$GDM is therefore structurally incompatible with our \pslr{} recipe (it violates C2) and is not assigned a discriminative score.

\paragraph{TMPD~\citep{boys2023tmpd}.}
Row-sum diagonal: $\mathbb{V}[x \mid x_t] \approx \mathrm{diag}(\mathbf{1}^\top J)$. Same fundamental issue as $\Pi$GDM: the row-sum collapses the structured block correlations created by an overcomplete lift.

\paragraph{DiffPIR~\citep{zhu2023diffpir}.}
Alternates denoising and proximal data-fidelity steps with no covariance. The proximal step requires solving $(I + \lambda G^\top G) x = \tilde{x} + \lambda G^\top y$; because $G^\top G = \tsimg \cdot \Ats^\top \cdot \Ats \cdot \imgts$ is not diagonal, this has no closed form, and inner solvers collapse DiffPIR to something similar to MMPS without the Bayesian justification.

\paragraph{MMPS~\citep{rozet2024diem}, our base sampler.}
Full Jacobian via Tweedie's covariance: $\mathbb{V}[x \mid x_t] \approx \sigma_t^2 J$ with $J = \nabla_{x_t} d_\theta$. The linear system $(\Sigma_y + \sigma_t^2 A J A^\top) v = r$ is solved via CG; each iteration costs one VJP. MMPS uses the full covariance and the CG solver handles arbitrary linear operators implicitly. However, MMPS was derived for single-space problems where $A$ acts in model space; in our dual-space setting it exhibits three structural misspecifications (App.~\ref{app:mmps-fails}) that the three \pslr{} corrections address.

\subsection{Classical EM convergence}
\label{app:em-convergence}

The expectation--maximization algorithm of~\citep{dempster1977em} fits a parameter $\theta$ of a generative model $p_\theta(x)$ from incomplete data $y$ by alternating two steps. Let $q$ be any distribution over the missing variable $\xts^{\mathrm{miss}}$ given $y$. Decompose the data log-likelihood as
\begin{equation}
\label{eq:free-energy}
\log p_\theta(y) \;=\; \mathcal{F}(q, \theta) \;+\; \mathrm{KL}\!\left(q\,\|\,p_\theta(\cdot \mid y)\right),
\end{equation}
where the \emph{free energy} (negative variational free energy) is $\mathcal{F}(q, \theta) = \mathbb{E}_q[\log p_\theta(\xts^{\mathrm{obs}}, \xts^{\mathrm{miss}})] + H(q)$. EM alternates:
\[
\textbf{E-step:}\quad q^{(k)} = \arg\max_q \mathcal{F}(q, \theta^{(k)}) = p_{\theta^{(k)}}(\cdot \mid y);
\quad
\textbf{M-step:}\quad \theta^{(k+1)} = \arg\max_\theta \mathcal{F}(q^{(k)}, \theta).
\]
Because $\mathrm{KL}(q^{(k)}\,\|\,p_{\theta^{(k)}}(\cdot \mid y)) = 0$ after the E-step and the M-step does not decrease $\mathcal{F}$, \eqref{eq:free-energy} immediately gives \emph{Neal--Hinton monotonicity}:
\begin{equation}
\log p_{\theta^{(k+1)}}(y) \;\geq\; \log p_{\theta^{(k)}}(y),
\end{equation}
with equality iff $(q^{(k)}, \theta^{(k)})$ is a fixed point~\citep{neal1998em,wu1983convergence}. This is the formal basis for the ``monotone improvement'' claim of the main paper, and it applies to the idealized loop with an exact E-step. Corollary~\ref{cor:approx-em} below states what holds for the implemented algorithm.

\paragraph{Monte-Carlo EM (MCEM).}
When the exact posterior $p_{\theta^{(k)}}(\cdot \mid y)$ is intractable, the standard approach \citep{wei1990mcem} replaces the expectation by a Monte-Carlo average over $S$ posterior samples $\{\xts^{\mathrm{miss},(s)}\}_{s=1}^S$:
\[
\hat{Q}(\theta \mid \theta^{(k)}) \;=\; \tfrac{1}{S}\sum_{s=1}^S \log p_\theta(\xts^{\mathrm{obs}}, \xts^{\mathrm{miss},(s)}).
\]
Monotonicity holds with high probability for $S$ large enough that the Monte-Carlo error is dominated by the deterministic improvement~\citep{wei1990mcem,booth1999maximizing}. Our co-evolving EM follows this scheme with $S = 1$ approximate posterior draw per observation, drawn by Algorithm~\ref{alg:estep}. With a single, approximate draw, the high-probability monotonicity of MCEM does not carry over. The three E-step corrections (C1--C3) and the STL warm start (App.~\ref{app:init}) are designed to reduce the E-step sampling error in the dual-space setting (App.~\ref{app:mmps-fails}), not to remove it. The corollary below makes the remaining effect of this error explicit.

\paragraph{Algorithm-level statement.} Let $q_k$ denote the distribution of the \pslr{} E-step samples at iteration $k$, and let $\varepsilon_k \triangleq \mathrm{KL}\big(q_k \,\|\, p_{\theta_k}(\cdot \mid y)\big) \ge 0$ be the E-step sampling error.

\begin{corollary}[Approximate monotonicity]\upshape
\label{cor:approx-em}
If the M-step does not decrease the free energy, $\mathcal{F}(q_k, \theta_{k+1}) \ge \mathcal{F}(q_k, \theta_k)$, then
\begin{equation*}
\log p_{\theta_{k+1}}(y) \;\ge\; \mathcal{F}(q_k, \theta_{k+1}) \;\ge\; \mathcal{F}(q_k, \theta_k) \;=\; \log p_{\theta_k}(y) - \varepsilon_k .
\end{equation*}
\end{corollary}
\begin{proof}
All three steps use the decomposition~\eqref{eq:free-energy} with $q = q_k$, the approximate sampler in place of the exact posterior. The first inequality is~\eqref{eq:free-energy} at $\theta_{k+1}$ together with the non-negativity of the KL term. The second is the assumption on the M-step. The equality is~\eqref{eq:free-energy} rearranged at $\theta_k$.
\end{proof}
\noindent With an exact E-step, $\varepsilon_k = 0$ and the corollary reduces to Neal--Hinton monotonicity. Its assumption is exactly the positivity of the M-step gap $\Delta Q$ measured in App.~\ref{app:em-diagnostics}, since the entropy term of $\mathcal{F}$ does not depend on $\theta$. That diagnostic is positive at every measured iteration ($20/20$).

\subsection{Why iteration: frozen-imputer bias and four mechanisms}
\label{app:why-close-full}

Open-loop pipelines train the generator on completions produced by a \emph{fixed} imputer; the four mechanisms below combine to give an irreducible bias relative to closed-loop EM. Each mechanism is shorthand for one term of the same distributional gap, formalised in Proposition~\ref{prop:frozen-bias}.

\paragraph{(i) Conditional-mean trap.}
\refstepcounter{appmech}\label{app:cond-mean-trap}
Deterministic imputers output point estimates, but generative models need distributional samples. Given $p(\xts^{\mathrm{miss}} \mid \xts^{\mathrm{obs}})$, the MSE-optimal estimate is the conditional mean. When the conditional is multi-modal, the conditional mean lies between modes in regions where the data has no support: a generator trained on mean-imputed data learns to generate at this empty point, which is wrong. \cref{tab:illustration-mse-trap} (App.~\ref{app:illustration}) visualises the resulting MSE/disc inversion.

\paragraph{(ii) Variance collapse.}
\refstepcounter{appmech}\label{app:var-collapse}
By the law of total variance,
\(
\mathrm{Var}(X) = \mathrm{Var}(\mathbb{E}[X \mid \mathrm{obs}]) + \mathbb{E}[\mathrm{Var}(X \mid \mathrm{obs})];
\)
replacing every imputed value by its conditional mean has variance $\mathrm{Var}(\mathbb{E}[X\mid\mathrm{obs}]) < \mathrm{Var}(X)$. A generator fit to the imputed dataset reproduces the collapsed variance directly. The deficit $\mathbb{E}[\mathrm{Var}(X\mid\mathrm{obs})]$ is largest when missingness is largest, exactly when the generator most needs distributional support~\citep{rubin1987multiple}.

\paragraph{(iii) Masking--data trade-off.}
\refstepcounter{appmech}\label{app:masking-tradeoff}
ImagenI2R~\citep{fadlon2025imageni2r} mitigates variance collapse by masking the loss to observed positions only. This avoids learning collapsed variance, but at missing rate $r$ only $(1-r)$ of pixels contribute to the loss, and noise is added everywhere at sampling time but only at observed positions during training, creating a train/test mismatch in the forward process. Closed-loop EM avoids both: after the E-step the data is complete, so the M-step uses the full loss with no masking and no mismatch.

\paragraph{(iv) Frozen-stochastic insufficiency.}
\refstepcounter{appmech}\label{app:frozen-stochastic}
Stochastic imputers (e.g.\ CSDI~\citep{tashiro2021csdi}) restore diversity but train once and stay fixed; their conditional distribution is set before the generator exists. Below we make this precise.

\begin{restatedprop}{prop:frozen-bias}
Assume standard regularity (smoothness; $\nabla_\theta$ commutes with $\mathbb{E}_{q_0}$; interior maxima). Let $q_0(\cdot \mid \xts^{\mathrm{obs}})$ be a fixed imputer, $\theta^\star \in \arg\max_\theta \log p_\theta(\xts^{\mathrm{obs}})$ a global MLE, and $\theta^\star_{q_0} \in \arg\max_\theta \mathcal{F}(q_0, \theta)$ a global maximizer of the open-loop M-step objective. Then for any $\theta$,
\begin{equation}
\nabla_\theta \log p_\theta(\xts^{\mathrm{obs}}) \;=\; \nabla_\theta \mathcal{F}(q_0, \theta) \;+\; \nabla_\theta \mathrm{KL}\!\left(q_0\,\|\,p_\theta(\cdot \mid \xts^{\mathrm{obs}})\right),
\end{equation}
and at $\theta^\star_{q_0}$ the first term vanishes:
\begin{equation}
\nabla_\theta \log p_\theta(\xts^{\mathrm{obs}}) \big|_{\theta^\star_{q_0}} \;=\; \nabla_\theta \mathrm{KL}\!\left(q_0\,\|\,p_\theta(\cdot \mid \xts^{\mathrm{obs}})\right) \big|_{\theta^\star_{q_0}}.
\end{equation}
Under EM self-consistency $q_0 = p_{\theta^\star_{q_0}}(\cdot \mid \xts^{\mathrm{obs}})$ this gradient also vanishes and $\theta^\star_{q_0}$ is a stationary point of $\log p_\theta(\xts^{\mathrm{obs}})$, an EM fixed point~\citep{wu1983convergence}; otherwise it is generically non-zero and $\theta^\star_{q_0}$ is strictly suboptimal. The open-loop gap to the MLE admits the upper bound
\begin{equation}
\label{eq:openloop-gap-upper}
0 \;\le\; \log p_{\theta^\star}(\xts^{\mathrm{obs}}) - \log p_{\theta^\star_{q_0}}(\xts^{\mathrm{obs}})
\;\le\;
\mathrm{KL}\!\left(q_0\,\|\,p_{\theta^\star}(\cdot \mid \xts^{\mathrm{obs}})\right)
\;-\;
\mathrm{KL}\!\left(q_0\,\|\,p_{\theta^\star_{q_0}}(\cdot \mid \xts^{\mathrm{obs}})\right).
\end{equation}
\end{restatedprop}

\begin{proof}
Differentiating the identity $\log p_\theta(\xts^{\mathrm{obs}}) = \mathcal{F}(q_0, \theta) + \mathrm{KL}(q_0\,\|\,p_\theta(\cdot|\xts^{\mathrm{obs}}))$ in $\theta$ (legal because $q_0$ is $\theta$-fixed) gives the gradient identity; the first-order condition $\nabla_\theta \mathcal{F}(q_0, \theta^\star_{q_0}) = 0$ kills the first term at $\theta^\star_{q_0}$. Under self-consistency, $\theta \mapsto \mathrm{KL}(q_0\,\|\,p_\theta(\cdot|\xts^{\mathrm{obs}}))$ attains its minimum value $0$ at $\theta^\star_{q_0}$, so its gradient vanishes. For \eqref{eq:openloop-gap-upper}: the lower bound is the definition of $\theta^\star$; for the upper bound, evaluate the identity at $\theta^\star$ and $\theta^\star_{q_0}$, subtract, and use $\mathcal{F}(q_0, \theta^\star_{q_0}) \ge \mathcal{F}(q_0, \theta^\star)$.
\end{proof}

The bias is governed by the \emph{distributional} object $\mathrm{KL}(q_0 \,\|\, p_\theta(\cdot \mid \xts^{\mathrm{obs}}))$. Mechanisms (i)--(iv) are consequences of the same gap: deterministic $q_0$ (Dirac) makes the KL infinite (mechanism (i)), while stochastic $q_0$ with the right moments makes it finite but generally non-zero because $q_0$ was built without reference to the generator (mechanism (iv)). A frozen $q_0$, chosen before training, has no way to satisfy the self-consistency condition with the not-yet-known $\theta^\star_{q_0}$; closed-loop EM is the construction that re-establishes it at every step via $q^{(k+1)} \leftarrow p_{\theta^{(k)}}(\cdot \mid \xts^{\mathrm{obs}})$, after which Neal--Hinton monotonicity (\ref{app:em-convergence}) guarantees $\log p_{\theta^{(k+1)}}(y) \ge \log p_{\theta^{(k)}}(y)$ in the idealized loop with an exact E-step. The proposition itself concerns the frozen imputer $q_0$ and does not depend on how accurate our E-step is.

\paragraph{Three predictions the proposition makes about the illustration.}
Proposition~\ref{prop:frozen-bias} predicts that (a) the frozen-stochastic baseline plateaus at a non-zero distance above the oracle, (b) pooling more samples cannot close it, and (c) one EM iteration is insufficient because $\theta^{(1)}$ is itself seeded from $q_0$. App.~\ref{app:illustration} verifies all three on a controlled $2$D experiment.

\paragraph{When is the target identifiable?}
The EM fixed point can only be the right target if the observed data determine the distribution we want to learn. Let $M \in \{0,1\}^{\Nts \times F}$ be the observation mask ($1$ = observed), and for a set of positions $S$ write $x_S$ for the values at $S$ and $M_S = \mathbf{1}$ for the event that all of them are observed. We assume
(A1) the mask is drawn independently of the values, $M \perp \xts$.

\begin{remark}[Identifiability under random masking]
\label{rem:identifiability}
Under (A1), every set of positions $S$ with $\Pr(M_S = \mathbf{1}) > 0$ satisfies $p(x_S) = p(x_S \mid M_S = \mathbf{1})$, and the right-hand side is a distribution of observed data. Hence the marginal $p(x_S)$ is determined by the distribution of the observations.
\end{remark}
\begin{proof}
By (A1), conditioning on any event that depends only on $M$ leaves the distribution of $\xts$, and hence of $x_S$, unchanged, so $p(x_S \mid M_S = \mathbf{1}) = p(x_S)$. Since $\Pr(M_S = \mathbf{1}) > 0$, this conditional distribution is well defined, and on that event every entry of $x_S$ is observed, so it is a distribution of observed values.
\end{proof}

\noindent What the Remark identifies depends on which sets are sometimes fully observed, which is a property of the data-collection protocol rather than of any method. (A1) holds by construction in all our benchmarks. Under per-cell masking, each entry is dropped independently with probability $\rho < 1$, so every set, including all positions, is fully observed with probability $(1-\rho)^{|S|} > 0$, and the full distribution $p(\xts)$ is identified. The standard benchmark instead drops a fixed fraction $r$ of the timesteps from every series, so no training series is ever fully observed. There, a set of positions is sometimes fully observed only if it spans at most $\lfloor (1-r)\Nts \rfloor$ timesteps, and the observations identify the marginals over any such set of timesteps, not the full joint distribution. Dependence across longer ranges is not identified by the observations alone. The continuous-time regime instead observes random linear combinations of the series, which falls under the random-measurement setting of~\citet{bora2018ambientgan}. Contiguous block gaps further limit which sets are ever observed together, and value-dependent missingness (App.~\ref{app:mnar}) breaks (A1). Our claim in those regimes is only that closing the loop gets measurably closer to the target than open-loop training (\cref{tab:avg-benchmarks}).

\subsection{Why single-space samplers fail in dual-space (M1, M2, M3)}
\label{app:mmps-fails}
\label{subsec:misspec}

Single-space Tweedie-moment samplers (DPS, $\Pi$GDM, TMPD, MMPS; App.~\ref{app:sampler-design-space}) are derived for $y = A\ximg + \mathrm{noise}$ with $A$ acting in the model's native space. In our dual-space setting the true operator is $G = \Ats \circ \imgts$, and dropping a single-space sampler in unchanged exhibits three structural misspecifications. We use MMPS as the running example: M1 and M3 concern the observation model and the lifted geometry and port verbatim to any Tweedie-moment sampler; M2's CG-conditioning argument has an analogue in TMPD's closed-form solve under the same adaptive-$\sigma_y$ fix (\cref{app:ltmpd} verifies the empirical transfer).

\paragraph{M1: wrong observation operator.}
\label{app:M1}
Standard MMPS uses an image-space mask $\Aimg \triangleq \mathrm{diag}(m_{\mathrm{img}})$ with $m_{\mathrm{img}} \triangleq \tsimg(m_{\mathrm{ts}})$. Because $\tsimg$ replicates each observed time point across overlapping windows, this lifted mask treats duplicated positions as independent observations: the CG system has dimension $\Nimg$ rather than the true $\Nobs \cdot F$ (e.g., $576$ vs.\ $12$ per channel on Energy with $24\!\times\!24$ images at $50\%$ missingness, a $48\times$ inflation). The score correction also points in the wrong direction: $J^\top \Aimg^\top r_{\mathrm{img}}$ acts on an image-space residual $r_{\mathrm{img}} = \Aimg(y_{\mathrm{img}} - \hat{x}_0)$, while the correct $J^\top G^\top r_{\mathrm{ts}}$ acts on $r_{\mathrm{ts}} = \yts - \Ats(\imgts(\hat{x}_0))$ and broadcasts back through $\tsimg \circ \Ats^\top$; these coincide only when $\tsimg$ is square and orthonormal, which delay embedding is not.

\paragraph{M2: ill-conditioned linear system.}
\label{app:M2}
MMPS solves $(\sigma_y^2 I + \sigma_t^2 A J A^\top)\,v = r$ at every reverse-diffusion step. In our noiseless setting $\Sigma_y = \sigma_y^2 I$ has no physical meaning (it is purely a regulariser), and two structural facts make any single fixed $\sigma_y$ pathological. First, with $A = \Aimg = \mathrm{diag}(m_{\mathrm{img}})$ (convention $m_i\!=\!1$ observed, $0$ missing), every missing position contributes a zero row and column to $\Aimg J \Aimg$, so $\lambda_{\min} = 0$ at all $\sigma_t$ and $\kappa = 1 + \sigma_t^2 \lambda_{\max}(\Aimg J \Aimg)/\sigma_y^2$ blows up for any small $\sigma_y$; CG within a few iterations fails to converge and returns null-space noise. Second, even with the corrected $G$, Tweedie's identity $\sigma_t^2 J = \mathbb{V}[x_0\!\mid\! x_t]$ implies the spectrum of $\sigma_t^2 G J G^\top$ saturates at $\mathbb{V}_{\mathrm{prior}}$ as $\sigma_t\!\to\!\infty$ and vanishes as $\sigma_t\!\to\!0$, so any fixed $\sigma_y$ is well-matched to at most a narrow band of noise levels. Single-space settings (CIFAR-$10$, MRI) include physical sensor noise ($\Sigma_y \sim 10^{-4} I$) that fortuitously matches the operating regime; in our noiseless TS setting we instead set $\sigma_y = c\,\sigma_t$ (correction C2 of \cref{subsec:pslr}), reducing the system to $\sigma_t^2(c^2 I + A J A^\top)\,v = r$, whose conditioning depends only on the data and on $c$, not on $\sigma_t$, so a single $c$ regularises every reverse step uniformly.

\paragraph{M3: no output-space constraint.}
\label{app:M3}
Valid images (those corresponding to actual time series) form a linear subspace $\mathrm{Range}(\tsimg) \subset \mathbb{R}^{\Nimg}$. MMPS outputs lie in all of $\mathbb{R}^{\Nimg}$ with no mechanism to constrain them. Off-manifold components compound across EM iterations: the M-step trains the denoiser to reproduce off-manifold structure; the next E-step's Tweedie mean inherits it; off-manifold energy accumulates. The ablation (\cref{tab:ablation}, row \emph{w/o C3}) confirms compounding catastrophically: stable CG produces accurate corrections in the wrong subspace, and the M-step faithfully learns this off-manifold structure. Correction C3 (manifold consistency) closes this loop using $\Pi = \tsimg \circ \imgts$ as the orthogonal projector.

\paragraph{One root cause, three consequences.}
M1, M2, M3 are three mathematical consequences of the single fact that MMPS was derived for $y = Ax + \mathrm{noise}$ where $A$ acts in model space. In dual-space the true model is $\yts = G \cdot \ximg$ where $G$ crosses spaces; the three \pslr{} corrections (C1, C2, C3) address each consequence in turn. The STL warm start of App.~\ref{app:init} is a complementary algorithmic choice that cuts the cold-start factor in early iterations but is not itself a posterior-sampling correction.

\begin{table}[ht]
\centering
\small
\setlength{\tabcolsep}{4pt}
\setlength{\aboverulesep}{1pt}
\setlength{\belowrulesep}{1.5pt}
\caption{Summary of the three failure modes. Each row gives a failure mode, the \pslr{} condition that fixes it, and the average Disc.\ score ($\downarrow$) when only that condition is removed. Scores are averaged over Energy, Weather and Stock at $30\%$, $50\%$ and $70\%$ missing rates (\cref{tab:ablation}). The full method scores $0.027$.}
\label{tab:failure-modes}
\begin{tabular}{l|c|c}
\toprule
\textbf{Failure mode} & \textbf{Fix} & \textbf{Disc.\ without fix} \\
\midrule
M1: conditions through an image-space mask instead of $G_i$ & C1 & 0.038 \\
M2: posterior uncertainty collapses, unstable linear solve & C2 & 0.047 \\
M3: samples drift off $\mathrm{Range}(\tsimg)$ across iterations & C3 & 0.056 \\
\bottomrule
\end{tabular}
\end{table}

\section{Controlled 2D Illustrative Experiment}
\label{app:illustration}

\paragraph{Setup.}
We complement the main ablation (\cref{tab:ablation}) with a simple controlled $2$D experiment where the true data distribution is fully known and can be visualized directly. The data consists of two interleaved spirals, and each coordinate is independently removed with probability $p_{\mathrm{miss}} = 0.60$.

We compare several methods that correspond directly to the rows in \cref{tab:ablation}. Most baselines first fill in the missing values and then train a generator on the completed data. In contrast, our co-evolving EM method alternates between sampling the missing values and re-training the generator, allowing both steps to improve together over multiple iterations.

We then evaluate how well the generated samples recover the original spiral structure using the discriminative score,
\[
\mathrm{disc} = |0.5 - \mathrm{accuracy}|,
\]
where lower values are better. A low score means the generated samples are difficult to distinguish from the true data. Full implementation details and hyperparameters are provided in \emph{Details}.

\begin{figure}[ht]
\centering
\includegraphics[width=\textwidth]{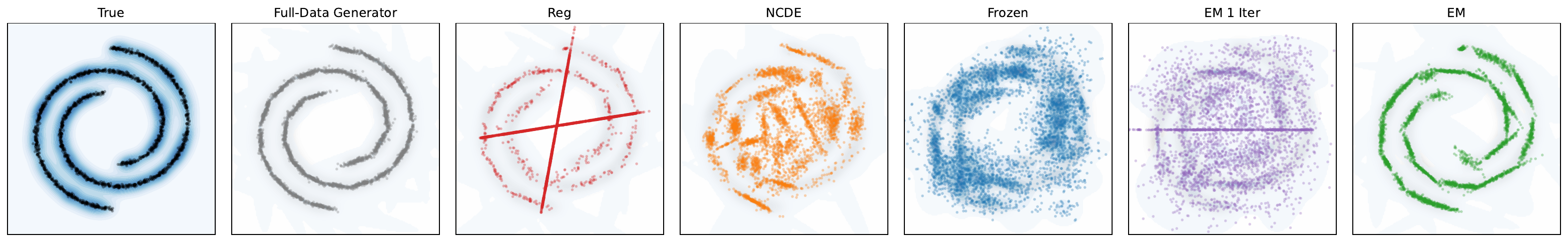}
\caption{Generated samples on two-spirals at $p_{\mathrm{miss}}{=}0.60$. }
\label{fig:illustration-samples}
\end{figure}

\paragraph{Methods.}
We compare five methods together with a \textbf{Full-data generator}, which is trained directly on the clean fully-observed data and serves as an upper-bound reference. The goal is to understand how different imputation strategies affect the quality of the final generated distribution. The results are shown in \cref{fig:illustration-samples}.

\textbf{Regression} \emph{(Reg in fig.)} fills each missing value using a linear regression model based on the observed coordinates. The completed dataset is then used to train a generator. Since the imputer predicts only a single average value, it tends to collapse the variability of the data.

\textbf{NCDE-style imputation} \emph{(NCDE in fig.)} follows the same pipeline, but replaces the linear regression with a more flexible non-linear imputer. This is the $2$D analogue of the imputation step used in GT-GAN / KoVAE-style methods.

\textbf{EM 1 Iter} \emph{(EM 1 iter in fig.)} replaces deterministic imputations with samples from the generator posterior, but performs only a single EM iteration. This corresponds to the \emph{Ours, $1$ iter} ablation in \cref{tab:ablation}.

\textbf{Frozen stochastic} \emph{(Frozen in fig.)} first trains a stochastic imputer until convergence, freezes it, and then trains the generator on samples produced by this fixed imputer. This is the $2$D analogue of CSDI-impute~\citep{tashiro2021csdi} followed by EDM training.

\textbf{Co-evolving EM (Ours)} \emph{(EM in fig.)} also uses stochastic imputations, but instead of freezing the imputer, it repeatedly alternates between re-sampling the missing values and re-training the generator. The imputer and generator therefore improve together over multiple iterations.

\paragraph{Main finding: better imputation error does not mean better generation.}
The method with the lowest imputation MSE, Point regression, produces the worst generated samples (\cref{fig:illustration-samples}). Instead of recovering the two spiral arms, it collapses the data into a regression-line cross. In contrast, Co-evolving EM achieves much better generation quality and successfully recovers both spirals, even though its imputation MSE is almost twice as large (\cref{tab:illustration-mse-trap}).

This shows that standard imputation metrics such as MSE and MAE can be misleading when the final goal is generation. These metrics reward methods that predict a single average value for every missing entry, even if this destroys the true structure and variability of the data.

In contrast, the Energy Score (ES)~\citep{gneiting2007strictly}, which evaluates the full conditional distribution rather than only a single point estimate, consistently agrees with the downstream generation metric (disc) across all configurations of the sweep.

\begin{table}[ht]
\centering
\caption{Comparison between point-estimate imputation metrics (MSE/MAE), the distributional Energy Score (ES), and downstream generation quality (disc) on the two-spirals dataset with $p_{\mathrm{miss}}{=}0.60$. Lower is better in all columns.}
\label{tab:illustration-mse-trap}
\resizebox{0.78\textwidth}{!}{
\begin{tabular}{l|cccc}
\toprule
\textbf{Method} & \textbf{imp.\ MSE} $\downarrow$ & \textbf{imp.\ MAE} $\downarrow$ & \textbf{imp.\ ES} $\downarrow$ & \textbf{disc} $\downarrow$ \\
\midrule
Point regression                & $\mathbf{2.22}$ & $\mathbf{1.31}$ & $1.31$          & $0.390$ \\
Point imputer (NCDE)            & $2.46$          & $1.28$           & $1.28$          & $0.252$ \\
EM Iter 1           & $3.65$          & $1.56$           & $0.91$          & $0.222$ \\
Frozen stochastic               & $4.31$          & $1.62$           & $0.85$          & $0.154$ \\
Co-evolving EM (Ours)           & $4.26$          & $1.47$           & $\mathbf{0.79}$ & $\mathbf{0.049}$ \\
\midrule
Full-data generator             & ---             & ---              & $0.72$          & $0.029$ \\
\bottomrule
\end{tabular}
}
\end{table}

\paragraph{Stochastic sampling helps, but iteration matters most.}
Moving from Point regression to Frozen stochastic greatly improves generation quality: the disc score drops from $0.390$ to $0.154$ on two-spirals at $p{=}0.60$. This shows that replacing single deterministic imputations with stochastic samples already preserves much more of the true data structure. However, stochastic sampling alone is not enough. The remaining gap is closed by iteration. Even the weak Stochastic ($1$ iter) version already performs similarly to Frozen stochastic, despite using only a single EM step. Co-evolving EM improves further by repeatedly alternating between re-sampling the missing values and re-training the generator, reaching a disc score of $0.049$, close to the Full-data generator baseline ($0.029$). Overall, the experiment shows that the main improvement comes not only from using stochastic imputations, but from allowing the imputer and generator to improve together over multiple iterations.

\paragraph{The improvement happens early.}
Co-evolving EM reaches its best performance after only a few EM iterations, achieving a disc score of $0.029 \pm 0.003$, close to the Full-data generator baseline ($0.026$, \cref{tab:illustration-robustness}). Most of the improvement appears in the first iterations: about half of the gain over Frozen stochastic is reached by iteration $4$, and about three quarters by iteration $7$. This mirrors the early-peak behavior observed in the main experiments (peak at $k{=}3$; Algorithm~\ref{alg:em-compact}).

\paragraph{Robustness across $21$ configurations.}
\label{app:illustration-robustness}
We repeated the experiment on $7$ different distributions (moons$+$ellipse, pinwheel-$5$, $9$-component grid, rings, checkerboard, two-spirals, and swirl) and $3$ missing rates ($0.50, 0.60, 0.70$), giving $21$ total configurations.

The overall ranking was consistent across the sweep:
\[
\text{Full-data generator}
<
\text{Co-evolving EM}
<
\text{Frozen stochastic}
<
\text{1 EM iter}
\]
\[
<
\text{Point imputer (NCDE)}
<
\text{Point regression}
\]
where lower disc is better. Median disc scores were $0.026$, $0.056$, $0.151$, $0.216$, $0.228$, and $0.373$, respectively. The early-improvement pattern of Co-evolving EM was also consistent across the sweep: performance improved monotonically until the peak in $20/21$ configurations, and the best result was always reached within at most $7$ EM iterations. \Cref{tab:illustration-robustness} shows that the pairwise comparisons hold almost universally across all configurations.

\begin{table}[ht]
\centering
\caption{Robustness across $21$ configurations of the controlled $2$D sweep.}
\label{tab:illustration-robustness}
\resizebox{0.82\textwidth}{!}{
\begin{tabular}{l|c}
\toprule
\textbf{Claim} & \textbf{Holds in} \\
\midrule
Co-evolving EM $<$ Frozen stochastic                                    & $\mathbf{21\,/\,21}$ \\
Frozen stochastic $<$ 1 EM iter                                         & $\mathbf{20\,/\,21}$ \\
Co-evolving EM $<$ Point imputer (NCDE)                                 & $\mathbf{21\,/\,21}$ \\
Co-evolving EM $<$ Point regression                                     & $\mathbf{21\,/\,21}$ \\
Co-evolving EM within $0.07$ disc of Full-data generator                & $\mathbf{21\,/\,21}$ \\
\bottomrule
\end{tabular}
}
\end{table}

\paragraph{Connection to the main ablation.}
\Cref{tab:illustration-correspondence} shows the direct correspondence between the controlled $2$D methods above and the methods used in the main ablation (\cref{tab:ablation}). The same ordering observed in the experiment also appears on the real datasets. On every one of the $21$ controlled configurations, iteration is the dominant lever between the best frozen-imputer baseline and near-oracle performance. This is the central claim of the paper: the main limitation is not the imputer itself, but whether imputations and generator are updated together iteratively. We show this on a fully visible controlled $2$D problem and observe the same behavior on real data in \cref{tab:ablation}. The controlled setup is intentionally operator-free, so the comparison is not affected by operator design choices. Operator-related failure modes are studied separately in App.~\ref{app:mmps-fails}; an equivalent controlled experiment cannot be performed for STFT representations (App.~\ref{app:stft}).

\begin{table}[ht]
\centering
\caption{Controlled-method $\leftrightarrow$ main-ablation correspondence. Each controlled baseline is a $1$-to-$1$ analog of one row of \cref{tab:ablation}; the controlled ordering predicts the real ordering in $5/5$ checked rows.}
\label{tab:illustration-correspondence}
\resizebox{0.95\textwidth}{!}{
\begin{tabular}{l|l|l}
\toprule
\textbf{Controlled method} & \textbf{Main-ablation analog} & \textbf{Prediction} \\
\midrule
Point regression       & ImagenTime, no masked-loss patch                        & worst disc / FID \\
Point imputer (NCDE)   & GT-GAN / deterministic NCDE imputation step             & deterministic ceiling, above Point regression \\
Stochastic ($1$ iter)  & Ours, $1$ EM iteration                                  & captures stochastic gain, misses iteration gain \\
Frozen stochastic      & CSDI-impute $+$ EDM (commit-and-forget)                 & best frozen baseline; bounded by capacity gap \\
Co-evolving EM (Ours)  & Full \pslr{}-EM ($K_{\mathrm{gen}}{=}30$, $12$ iters)   & oracle-matching; wins in $21/21$ \\
\bottomrule
\end{tabular}
}
\end{table}

\paragraph{Details.}
{\small The data consists of two interleaved spirals with Gaussian noise $\sigma{=}0.12$. We use Gaussian mixture models (GMMs) throughout the experiment, with $K_{\mathrm{imp}}{=}3$ mixture components for the imputer and $K_{\mathrm{gen}}{=}30$ mixture components for the generator. We use $N_{\mathrm{train}}{=}2000$, $N_{\mathrm{eval}}{=}1500$, $N_{\mathrm{reps}}{=}5$, and random seed $42$. Co-evolving EM and Frozen stochastic use $12$ iterations, while 1 EM iter uses a single iteration. For \cref{tab:illustration-mse-trap}, stochastic methods use $M{=}50$ posterior samples when computing the Energy Score.

The disc score is computed using a Gradient Boosting classifier. Since the data is simple $2$D tabular data rather than time series, we use a standard tabular classifier instead of the RNN metric from App.~\ref{app:metrics}; the definition $\mathrm{disc}=|0.5-\mathrm{accuracy}|$ remains unchanged.

The Energy Score (ES)~\citep{gneiting2007strictly} measures how well a method matches the full conditional distribution rather than only a single prediction:
\[
\mathrm{ES}(\hat F, x)
=
\mathbb{E}\|\hat X - x\|
-
\frac12 \mathbb{E}\|\hat X - \hat X'\|.
\]
Unlike MSE and MAE, which reward single average predictions, ES rewards preserving the full variability and multimodal structure of the data.}

\section{Experimental Setup}
\label{app:setup-full}

This section gives the full experimental protocol used in the main paper: datasets (App.~\ref{app:datasets}), baselines (App.~\ref{app:baselines}), evaluation metrics with formulas (App.~\ref{app:metrics}), and per-dataset hyperparameters (App.~\ref{app:hyperparams}).

\subsection{Datasets}
\label{app:datasets}

We share the dataset suite with ImagenI2R~\citep{fadlon2025imageni2r} for direct comparability: one synthetic and several real-world multivariate time-series benchmarks spanning smooth periodicity, non-stationary financial random walks, high-dimensional physics, noisy-periodic energy, and meteorological fields. All main-table experiments use sequence length $24$; extended sequence lengths ($96$ and $768$) are reported in App.~\ref{app:full-results}.

\paragraph{Sine, Stocks, MuJoCo.} Sine is a synthetic $5$-feature dataset with $x_i(t) = \sin(2\pi f_i t + \theta_i)$, $f_i \sim \mathcal{U}[0,1]$, $\theta_i \sim \mathcal{U}[-\pi, \pi]$. Stocks is daily Google stock data ($2004$--$2019$, $6$ features: high, low, open, close, adjusted close, volume); it lacks periodicity and is dominated by random walks. MuJoCo (Multi-Joint dynamics with Contact)~\citep{todorov2012mujoco} provides $14$-feature physics rollouts and stresses smooth-but-high-dimensional dynamical structure.

\paragraph{Energy, Weather.} Energy~\citep{candanedo2017data} is the UCI Appliance Energy Prediction benchmark, with $28$ correlated features (appliance load, temperature/humidity from $9$ home zones, weather) sampled every $10$ min; at $\Nts{=}24$ each window covers $4$ h. Weather is the Max-Planck meteorological benchmark with $21$ features (temperature, humidity, pressure, wind, radiation, precipitation) at the same sampling rate. Energy is our primary benchmark because its high feature count, non-trivial cross-correlations, and realistic observation noise make it the most discriminative for revealing the masking--data trade-off and the conditional-mean trap.

\paragraph{ETTh1/2 and ETTm1/2.} The ETTh datasets~\citep{zhou2021informer} record electricity load and oil temperature from two power-station transformers ($7$ features each) sampled hourly; ETTm samples the same physical processes every $15$ min. ETTh stresses our method on low-sampling-frequency regimes where the delay-embedding image is sparse along the column axis; ETTm is the dense-sampling counterpart with larger adjacent-column correlation in the image (smaller $\|\Delta\|$, faster regime-(a) contraction).

\paragraph{Electricity.} Electricity~\citep{misc_electricityloaddiagrams20112014_321} is hourly consumption from $370$ clients (we subsample to the standard $21$-client feature set); strong diurnal periodicity makes it the cleanest test for the STL warm start.

\subsection{Baselines}
\label{app:baselines}

\paragraph{Published open-loop baselines.}
\textbf{GT-GAN}~\citep{jeon2022gtgan}: NCDE one-shot imputation $\to$ GAN. \textbf{KoVAE}~\citep{naiman2024kovae}: NCDE interpolation $\to$ Koopman VAE; strong VAE-based baseline. \textbf{ImagenI2R}~\citep{fadlon2025imageni2r}: TST-based completion $\to$ masked-loss diffusion; the strongest published baseline on this benchmark. \textbf{TimeGAN-$\Delta t$}~\citep{yoon2019timegan,jeon2022gtgan,che2018recurrent}: end-to-end RNN with $\Delta t$ markers (architectural family).

\paragraph{Stochastic-imputer steelman (ours, diagnostic).}
\textbf{CSDI-impute $+$ EDM}: train CSDI~\citep{tashiro2021csdi} for $200$ epochs on the irregular data, sample stochastic completions, then train a vanilla EDM~\citep{karras2022edm} on those completions. This is constructed solely to sharpen the EM argument: it isolates ``deterministic vs.\ stochastic imputer'' from ``open vs.\ closed loop''.

\paragraph{Alternative E-step posterior sampler.}
\textbf{Lifted-mask MMPS-EM} uses MMPS in our EM loop without any of the three \pslr{} corrections: it violates C1 (operator-consistent conditioning) by using the image-space mask $\Aimg$ in place of the composed operator $G_i$, violates C2 (uncertainty preservation) by using a fixed $\sigma_y = 0.01$ instead of the adaptive $\sigma_y = c\,\sigma_t$ schedule, and violates C3 (manifold consistency) by omitting the projection $\Pi = \tsimg \circ \imgts$; it also forgoes the STL warm start, i.e.\ MMPS as derived for single-space settings, dropped into the lifted setting unchanged. We do not include DPS-EM, $\Pi$GDM-EM, or DiffPIR-EM as iterative baselines: DPS and $\Pi$GDM violate C2 (full Tweedie covariance) and DiffPIR's proximal step admits no closed form once $A$ is composed with $\imgts$, so none of them can be slotted into Algorithm~\ref{alg:em-compact} as a faithful E-step (App.~\ref{app:mmps-fails}). They are therefore not assigned a discriminative score.

\paragraph{Ours.}
\textbf{Ours ($1$ iter)} runs the full PSLR recipe but stops after one EM iteration (a stochastic commit-and-forget analog isolating the pure-iteration effect). \textbf{Ours} (full) uses all three \pslr{} corrections plus the STL warm start and runs $K = 5$ EM iterations (peak at $k = 3$).

\subsection{Metrics}
\label{app:metrics}

We follow the standard time-series-generation evaluation protocol \citep{yoon2019timegan, fadlon2025imageni2r}, using four complementary quantitative metrics that together cover \emph{distributional indistinguishability} (discriminative score), \emph{downstream utility} (predictive score), \emph{representation-level fidelity} (Context-FID), and \emph{cross-feature dependency preservation} (correlational score), plus two qualitative checks (t-SNE and PDF overlap; \cref{sec:qualitative}). Lower is better for all four quantitative metrics. All numbers in this paper are averaged over $10$ independent runs with fresh seeds.

\paragraph{Discriminative score (primary)~\citep{yoon2019timegan}.}
A model-free, post-hoc test of whether real and synthetic samples are distributionally indistinguishable. We train a two-layer GRU binary classifier on a balanced mix of real and generated sequences to predict the source label, and report $|0.5 - \mathrm{accuracy}|$ on a held-out split. A score of $0$ means the classifier cannot do better than chance (synthetic is indistinguishable from real); $0.5$ means perfect separation. We use the discriminative score as the primary headline metric across all benchmark tables because it directly probes the property our method targets, closing the distributional gap between $p_\theta$ and the true data distribution, without committing to any specific downstream task.

\paragraph{Predictive score~\citep{yoon2019timegan}.}
A downstream-utility test under the ``train on synthetic, test on real'' (TSTR) protocol. We train a one-step-ahead GRU regressor using only the generated sequences as training data, and evaluate its mean-absolute-error (MAE) on the real test set. Low predictive score implies the synthetic sequences carry the temporal dynamics needed for next-step prediction, going beyond the marginal-fidelity probed by the discriminative score: a model that matches one-time-step marginals but corrupts the dependency between consecutive steps will pass discriminative-score scrutiny but fail TSTR.

\paragraph{Context-FID~\citep{jeha2022psafid,yue2022ts2vec}.}
A time-series adaptation of the Fr\'{e}chet Inception Distance. The original FID encodes images with a pretrained Inception network; \citet{jeha2022psafid} replace this encoder with TS$2$Vec~\citep{yue2022ts2vec}, a contrastive representation-learning model trained to produce per-timestep embeddings that respect the local temporal context of a sequence. Both real and generated samples are encoded by TS$2$Vec, and the standard Fr\'{e}chet distance between the two embedding distributions is reported. \citet{jeha2022psafid} show that lower Context-FID strongly correlates with better downstream forecasting and is more sensitive than discriminative score in regimes where samples are visually plausible but distributionally biased.

\paragraph{Correlational score~\citep{liao2020conditional}.}
A test of whether the cross-feature dependency structure of the data is preserved. For a sequence of $T$ time-steps with features indexed by $i, j$, we estimate the empirical covariance per pair as
\[
\mathrm{cov}_{i,j} \;=\; \tfrac{1}{T} \sum_{t=1}^{T} X_t^i X_t^j \;-\; \bigl(\tfrac{1}{T} \sum_{t=1}^{T} X_t^i\bigr)\bigl(\tfrac{1}{T} \sum_{t=1}^{T} X_t^j\bigr) \ ,
\]
and report the normalised $\ell_1$ gap between the real ($r$) and synthetic ($f$) Pearson-correlation matrices,
\[
\tfrac{1}{10} \sum_{i,j} \left| \tfrac{\mathrm{cov}_r^{i,j}}{\sqrt{\mathrm{cov}_r^{i,i}\,\mathrm{cov}_r^{j,j}}} \;-\; \tfrac{\mathrm{cov}_f^{i,j}}{\sqrt{\mathrm{cov}_f^{i,i}\,\mathrm{cov}_f^{j,j}}} \right| \ .
\]
A model can match per-feature marginals while distorting cross-feature relationships, e.g.\ training on point-imputed completions tends to over-smooth co-variation; a low correlational score certifies that the multivariate dependency structure has survived training.

\paragraph{Qualitative metrics: t-SNE and PDF overlap.}
For visual sanity-checks (\cref{sec:qual_eval_full}), we follow \citep{yoon2019timegan, fadlon2025imageni2r} and use two standard projections of the multivariate sequences. (i) \emph{t-SNE}~\citep{van2008visualizing} embeds real and synthetic sequences into a shared $2$D space; tight overlap of the two clouds is a positive indicator of distributional alignment. (ii) \emph{PDF overlap}: kernel-density-estimated marginal distributions of real and synthetic sequences are plotted on a common axis; alignment of mean and tails is a positive indicator of single-feature fidelity. These visualisations are not used to rank methods (the four quantitative metrics above do that) but they provide an at-a-glance failure check: e.g., variance-collapse from frozen point imputers shows up as a visibly narrower synthetic PDF, even when the discriminative score is moderate.

\subsection{Hyperparameters}
\label{app:hyperparams}

The \emph{algorithmic} hyperparameters of our method are fixed globally (App.~\ref{app:algorithms}); the \emph{architectural} hyperparameters vary per dataset and are reported in \cref{tab:hp_seq_24,tab:hp_seq_96,tab:hp_seq_768}.

\section{Additional Experiments}
\label{app:additional-exp}

This appendix provides the per-dataset, per-metric, per-rate breakdown behind the averaged numbers of~\cref{tab:avg-benchmarks} (main paper, \cref{sec:results}) and the extended qualitative comparison behind~\cref{fig:70_tsne_density}. The structure follows the Results section one-to-one: \cref{app:full-results} expands the \emph{Standard} benchmark (\cref{sec:quant-standard}), \cref{app:full-block} the three \emph{Block} positions (\cref{sec:quant-block}), \cref{app:full-cell} the \emph{Per-cell} asynchronous setting (\cref{sec:quant-cell}), and \cref{app:full-continuous} the \emph{Continuous-time} setting (\cref{sec:quant-continuous}). \cref{app:qualitative-wass} extends the $70\%$-rate snapshot of \cref{sec:qualitative} with the full t-SNE / KDE comparison at $30\%$, $50\%$, and $70\%$ together with the Wasserstein-distance table. We further validate the generality of the framework in \cref{app:stft,app:ltmpd}, showing that the \pslr{} recipe transfers without retuning both across lift representations (STFT) and across posterior samplers (\ltmpd{}).

All tables compare against the same four baselines fixed in the main paper (TimeGAN-$\Delta t$~\citep{yoon2019timegan,che2018recurrent}, GT-GAN~\citep{jeon2022gtgan}, KoVAE~\citep{naiman2024kovae}, and ImagenI2R~\citep{fadlon2025imageni2r}), and report all four metrics (discriminative, predictive, Context-FID, and correlational; lower is better throughout). Numbers for our method are the minimum across the EM training history.

\begin{table}[ht]
\centering
\caption{Architectural hyperparameters at sequence length $24$. Algorithmic knobs ($c$, $n_{\mathrm{CG}}$, $K$, $p_{\mathrm{proj}}$, STL) are global and not dataset-specific.}
\label{tab:hp_seq_24}
\setlength{\tabcolsep}{4pt}
\resizebox{\textwidth}{!}{
\begin{tabular}{l|cccccccccc}
\toprule
& \textbf{Sine} & \textbf{Stocks} & \textbf{MuJoCo} & \textbf{Energy} & \textbf{Weather} & \textbf{ETTh1} & \textbf{ETTh2} & \textbf{ETTm1} & \textbf{ETTm2} & \textbf{Electricity} \\
\midrule
Image size              & $8\times 8$ & $8\times 8$ & $8\times 8$ & $8\times 8$ & $8\times 8$ & $8\times 8$ & $8\times 8$ & $8\times 8$ & $8\times 8$ & $8\times 8$ \\
Delay $n$               & $8$ & $8$ & $8$ & $8$ & $8$ & $8$ & $8$ & $8$ & $8$ & $8$ \\
Skip $m$                & $3$ & $3$ & $3$ & $3$ & $3$ & $3$ & $3$ & $3$ & $3$ & $3$ \\
U-Net base channels     & $128$ & $128$ & $64$ & $128$ & $128$ & $128$ & $128$ & $128$ & $128$ & $128$ \\
Batch size              & $128$ & $128$ & $128$ & $128$ & $128$ & $128$ & $128$ & $128$ & $128$ & $128$ \\
Learning rate           & $1\!\cdot\!10^{-4}$ & $1\!\cdot\!10^{-4}$ & $1\!\cdot\!10^{-4}$ & $1\!\cdot\!10^{-4}$ & $1\!\cdot\!10^{-4}$ & $1\!\cdot\!10^{-4}$ & $1\!\cdot\!10^{-4}$ & $1\!\cdot\!10^{-4}$ & $1\!\cdot\!10^{-4}$ & $1\!\cdot\!10^{-4}$ \\
M-step epochs / iter    & $40$ & $40$ & $40$ & $40$ & $40$ & $40$ & $40$ & $40$ & $40$ & $40$ \\
\bottomrule
\end{tabular}
}
\end{table}

\begin{table}[ht]
\centering
\caption{Architectural hyperparameters at sequence length $96$. Algorithmic knobs ($c$, $n_{\mathrm{CG}}$, $K$, $p_{\mathrm{proj}}$, STL) are global and not dataset-specific.}
\label{tab:hp_seq_96}
\setlength{\tabcolsep}{4pt}
\resizebox{\textwidth}{!}{
\begin{tabular}{l|cccccccc}
\toprule
& \textbf{Sine} & \textbf{Stocks} & \textbf{Energy} & \textbf{Weather} & \textbf{ETTh1} & \textbf{ETTh2} & \textbf{ETTm1} & \textbf{ETTm2} \\
\midrule
Image size              & $16\times 16$ & $16\times 16$ & $32\times 32$ & $32\times 32$ & $16\times 16$ & $16\times 16$ & $16\times 16$ & $16\times 16$ \\
Delay $n$               & $16$ & $16$ & $32$ & $32$ & $16$ & $16$ & $16$ & $16$ \\
Skip $m$                & $6$ & $6$ & $24$ & $24$ & $6$ & $6$ & $6$ & $6$ \\
U-Net base channels     & $128$ & $128$ & $128$ & $128$ & $128$ & $128$ & $128$ & $128$ \\
Batch size              & $128$ & $128$ & $16$ & $32$ & $32$ & $64$ & $128$ & $128$ \\
Learning rate           & $1\!\cdot\!10^{-4}$ & $1\!\cdot\!10^{-4}$ & $1\!\cdot\!10^{-4}$ & $1\!\cdot\!10^{-4}$ & $1\!\cdot\!10^{-4}$ & $1\!\cdot\!10^{-4}$ & $1\!\cdot\!10^{-4}$ & $1\!\cdot\!10^{-4}$ \\
M-step epochs / iter    & $40$ & $40$ & $40$ & $40$ & $40$ & $40$ & $40$ & $40$ \\
\bottomrule
\end{tabular}
}
\end{table}

\begin{table}[ht]
\centering
\caption{Architectural hyperparameters at sequence length $768$. The image is kept fixed at $32 \times 32$ via skip $m{=}24$; one image therefore encodes $\sim 768$ time-series points. Algorithmic knobs ($c$, $n_{\mathrm{CG}}$, $K$, $p_{\mathrm{proj}}$, STL) are global and not dataset-specific.}
\label{tab:hp_seq_768}
\setlength{\tabcolsep}{4pt}
\resizebox{\textwidth}{!}{
\begin{tabular}{l|cccccccc}
\toprule
& \textbf{Sine} & \textbf{Stocks} & \textbf{Energy} & \textbf{Weather} & \textbf{ETTh1} & \textbf{ETTh2} & \textbf{ETTm1} & \textbf{ETTm2} \\
\midrule
Image size              & $32\times 32$ & $32\times 32$ & $32\times 32$ & $32\times 32$ & $32\times 32$ & $32\times 32$ & $32\times 32$ & $32\times 32$ \\
Delay $n$               & $32$ & $32$ & $32$ & $32$ & $32$ & $32$ & $32$ & $32$ \\
Skip $m$                & $24$ & $24$ & $24$ & $24$ & $24$ & $24$ & $24$ & $24$ \\
U-Net base channels     & $128$ & $128$ & $128$ & $128$ & $128$ & $128$ & $128$ & $128$ \\
Batch size              & $32$ & $32$ & $16$ & $32$ & $32$ & $32$ & $32$ & $32$ \\
Learning rate           & $1\!\cdot\!10^{-4}$ & $1\!\cdot\!10^{-4}$ & $1\!\cdot\!10^{-4}$ & $1\!\cdot\!10^{-4}$ & $1\!\cdot\!10^{-4}$ & $1\!\cdot\!10^{-4}$ & $1\!\cdot\!10^{-4}$ & $1\!\cdot\!10^{-4}$ \\
M-step epochs / iter    & $40$ & $40$ & $40$ & $40$ & $40$ & $40$ & $40$ & $40$ \\
\bottomrule
\end{tabular}
}
\end{table}

\subsection{Standard benchmark: per-dataset results}
\label{app:full-results}

The Standard benchmark (\cref{sec:quant-standard}) drops timesteps synchronously across features at rates $\{30\%, 50\%, 70\%\}$, following~\citep{yoon2019timegan,jeon2022gtgan,fadlon2025imageni2r}. \cref{tab:full_irregular_24,tab:irregular_96,tab:irregular_768} expand the \emph{Standard} column of \cref{tab:avg-benchmarks} into the four-metric, per-dataset view at sequence lengths $24$/$96$/$768$: length $24$ covers all ten datasets; lengths $96$ and $768$ follow the ImagenI2R protocol on eight (Mujoco and Electricity omitted). Across the resulting $78$ (dataset, rate, length) cells, Ours is the strictly best on Disc in $63$: $20/30$ at length $24$, $19/24$ at $96$, and a clean $24/24$ at $768$. The gap widens monotonically with both length and missing rate-the regime in which the conditional-mean trap (App.~\ref{app:cond-mean-trap}) is sharpest-and peaks at $70\%$ on the high-dimensional, non-periodic sets (Weather: $0.029$ vs.\ ImagenI2R $0.106$; Energy: $0.072$ vs.\ $0.128$; Etth1: $0.026$ vs.\ $0.058$). The cells where ImagenI2R remains competitive are smooth low-dimensional Sine and Mujoco at $30\%$, where conditionals are near-unimodal and a single point estimate is already close to a posterior sample; closing the loop has nothing to add there, by design.

\subsection{Block-structured missingness: per-dataset results}
\label{app:full-block}

Block-structured missingness (\cref{sec:quant-block}) removes a contiguous segment either from the beginning, a random location, or the end of the sequence, stressing warm-start dependence, interpolation, and extrapolation, respectively. \cref{tab:block_start_24,tab:block_random_24,tab:block_end_24} expand the three \emph{Block} columns of \cref{tab:avg-benchmarks} into per-dataset, four-metric views at length $24$.

As the corruption ratio increases, the open-loop pipelines degrade sharply, especially when the missing block appears at the end of the sequence. With $70\%$ block-end corruption, ImagenI2R reaches near-random discriminator scores on all ETT datasets ($0.498$--$0.500$), while KoVAE approaches collapse on Weather and Energy ($0.481/0.487$). In contrast, Ours stays below $0.103$ Disc. on all ETT datasets, yielding relative improvements of approximately $79\%$--$96\%$ over the strongest open-loop baselines. Under block-random corruption with $70\%$ missingness, Ours also improves Disc. by up to $\sim 91\%$ compared to ImagenI2R.

The improvements are particularly large on the harder datasets. Under $70\%$ block-start corruption, Ours reduces Disc. from $0.500$ to $0.126$ on Weather ($74.8\%$ improvement) and from $0.496$ to $0.426$ on Energy ($14.1\%$ improvement). Similar behavior appears across FID and Corr., where the gains often exceed one order of magnitude relative to the open-loop pipelines.

Under such heavy block corruption, the contiguous gap removes structure that the observations cannot uniquely identify, particularly in the block-start variant, where the model has $70\%$ of the sequence to predict before any context arrives. The learned model can therefore fit the observed prefix/suffix accurately while still differing from the truth on the missing block; this is the identifiability ceiling already flagged in the Limitations of \cref{sec:conclusions}, and explains why the residual gap on Energy block-start at $70\%$ ($0.426$) is operator-driven rather than loop-design-driven.

\subsection{Per-cell asynchronous missingness: per-dataset results}
\label{app:full-cell}

Per-cell missingness (\cref{sec:quant-cell}) drops each $(t,f)$ entry independently, producing an asynchronous observation mask across features. \cref{tab:cell_random_24_fixed_final} expands the \emph{Per-cell} column of \cref{tab:avg-benchmarks} into a per-dataset, four-metric view at length $24$. This setting mainly changes the corruption granularity relative to the Standard setting, isolating the effect of the masking pattern itself.

At $30\%$ and $50\%$ missingness, Ours achieves the best Disc. score on every dataset. The relative improvements are particularly large on Weather and Energy, where at $50\%$ missingness Ours reduces Disc. from $0.498$ to $0.021$ on Weather ($95.8\%$ improvement) and from $0.500$ to $0.114$ on Energy ($77.2\%$ improvement). Similar trends appear across FID and Corr., where Ours consistently outperforms the open-loop pipelines, often by one order of magnitude or more.

At $70\%$ missingness, the setting becomes substantially harder on the higher-dimensional datasets (Ettm1, Weather, and Energy), where removing $70\%$ of the full feature-time grid leaves very limited local structure for interpolation. Accordingly, Disc. rises to $0.458/0.450/0.449$ on these datasets. Still, Ours maintains a clear margin over the near-random baseline range ($0.475$--$0.500$). On the lower-dimensional datasets, performance remains stable even at $70\%$ missingness, with Disc. staying at $0.025$ on Etth1 and $0.007$ on Stock. This suggests that the degradation mainly reflects the reduced data-density regime rather than a limitation of the closed-loop training itself.

\subsection{Continuous-time irregular sampling: per-dataset results}
\label{app:full-continuous}

Real sensors rarely tick on a perfect clock: instead of recording at $t = 0, 1, 2, \dots$ they fire at irregular times like $t = 0.3, 1.7, 3.25, \dots$. To simulate this, we start from a clean sequence $x \in \mathbb{R}^{T \times F}$ on the integer grid $t = 0, \dots, T{-}1$ and draw $N_{\mathrm{obs}} = \mathrm{round}(\rho T)$ real-valued timestamps uniformly on $[0, T{-}1]$, independently per sample and sorted; here $\rho$ is the \emph{observation density} ($\rho{=}0.5$ keeps half the grid points, $\rho{=}0.3$ keeps $30\%$). The value at each timestamp $t_k$ is the linear blend of the two nearest integer-grid points, weighted by closeness:
\[
y_k = (1-\alpha)\, x_{\lfloor t_k \rfloor} + \alpha\, x_{\lfloor t_k \rfloor + 1}, \qquad \alpha = t_k - \lfloor t_k \rfloor.
\]
For example, $t_k{=}3.25$ sits a quarter of the way from $x_3$ to $x_4$, so the closer point ($x_3$) gets the larger weight: $y_k = 0.75\, x_3 + 0.25\, x_4$. Likewise $t_k{=}5.6$ yields $y_k = 0.4\, x_5 + 0.6\, x_6$, and an integer $t_k$ degenerates to the standard timestep mask (a single grid value contributes, $\alpha{=}0$). The model therefore observes only a sorted set of $\{(t_k, y_k)\}_{k=1}^{N_{\mathrm{obs}}}$ pairs at irregular times, never the underlying grid values. During training it learns from these irregular observations; at evaluation it generates full sequences back on the original integer grid. We evaluate on ETTh1, Energy, Weather, and Stock.

The full \pslr{} recipe applies unchanged: only the time-series operator $\Ats$ inside the composed operator $G_i = \Ats^{(i)} \circ \imgts$ (operator-consistent conditioning, C1) is replaced. The lift, the prior, uncertainty preservation (C2, including the adaptive $\sigma_y = c\,\sigma_t$ schedule), and manifold consistency (C3, enforced by projection with $\Pi = \tsimg \circ \imgts$) are kept fixed. This is the ``new $\Ats$'' axis of the modularity claim of \cref{subsec:algorithm}.

\paragraph{Adapting prior baselines to continuous-time inputs.} The four prior baselines assume grid-aligned observations and have no native handling of fractional timestamps, so they cannot be run on the continuous-time inputs as-is. To run them at all, we project each continuous observation onto a regular grid: pick a fixed sampling interval matching the finest granularity of interest for the application (e.g., $1$ second, $1$ minute), redefine the time axis as evenly spaced cells of that interval, snap each observation to its nearest cell, and mark every cell that received no observation as missing (NaN). For instance, observations at $t = 1, 4, 6$ on a unit-interval grid become the grid $t = 1, 2, 3, 4, 5, 6$ with values placed at $t = 1, 4, 6$ and NaN at $t = 2, 3, 5$. This surrogate is fundamentally limited: the smaller the chosen interval, the more cells end up empty, so the corruption pattern grows arbitrarily harder for any method that does not natively model continuous time.

\cref{tab:continuous_24_filled} expands the \emph{Continuous} column of \cref{tab:avg-benchmarks} into the four-metric, per-dataset view at length $24$ for $\rho \in \{0.3, 0.5\}$. The meaningful contrast is therefore against the clean-data oracle: at $\rho{=}0.5$ Ours' Disc is within a small additive gap on ETTh1 ($0.019$ vs.\ $0.006$), Weather ($0.021$ vs.\ $0.003$), and Energy ($0.071$ vs.\ $0.009$), and matches the oracle on Stock ($0.006$ vs.\ $0.010$); the same pattern holds at $\rho{=}0.3$ on three of four sets. The Energy outlier at $\rho{=}0.3$ ($0.180$ vs.\ $0.009$) is the lone cell where the high feature count ($28$) interacts adversely with the linear-blend operator: each fractional observation entangles two adjacent grid rows across all $28$ channels, and at $\rho{=}0.3$ the $\sim\!7$ surviving observations per channel are too sparse to disentangle; raising $\rho$ to $0.5$ (a $\sim\!1.7\times$ budget increase) closes the gap, consistent with an operator-budget rather than loop-design diagnosis.

\begin{table}[H]
\centering
\small
\setlength{\tabcolsep}{3pt}
\caption{\textbf{Wasserstein distances} (Real vs.\ generated) for the qualitative figures above. $\mathcal{W}_2$ on the joint 2-D t-SNE of per-window means (ImagenI2R protocol~\citep{fadlon2025imageni2r}); $\mathcal{W}_1$ on the per-window marginal shown by the bottom-row KDEs. Lower is better; \textbf{bold} marks the lower number per (metric, dataset). Ours wins every cell on both metrics.}
\label{tab:wass_dist}
\resizebox{\columnwidth}{!}{
\begin{tabular}{cl|ccc|ccc|ccc}
\toprule
& & \multicolumn{3}{c|}{\textbf{$30\%$ Drop}} & \multicolumn{3}{c|}{\textbf{$50\%$ Drop}} & \multicolumn{3}{c}{\textbf{$70\%$ Drop}} \\
\textbf{Metric} & \textbf{Model}
& ETTh1 & ETTm2 & Energy
& Weather & ETTh2 & ETTm1
& Weather & Energy & Stocks \\
\midrule
\multirow{2}{*}{$\mathcal{W}_2$ t-SNE $\downarrow$}
& ImagenI2R     & $2.85$ & $1.88$ & $1.88$ & $1.44$ & $1.82$ & $1.82$ & $1.64$ & $1.78$ & $2.93$ \\
& \textbf{Ours} & \cellcolor{blue!10}$\boldsymbol{1.12}$ & \cellcolor{blue!10}$\boldsymbol{1.04}$ & \cellcolor{blue!10}$\boldsymbol{0.82}$ & \cellcolor{blue!10}$\boldsymbol{0.95}$ & \cellcolor{blue!10}$\boldsymbol{0.99}$ & \cellcolor{blue!10}$\boldsymbol{1.03}$ & \cellcolor{blue!10}$\boldsymbol{1.31}$ & \cellcolor{blue!10}$\boldsymbol{0.92}$ & \cellcolor{blue!10}$\boldsymbol{1.06}$ \\
\midrule
\multirow{2}{*}{$\mathcal{W}_1$ KDE $\downarrow$ ($\times 10^{-3}$)}
& ImagenI2R     & $6.2$ & $6.7$ & $5.2$ & $4.7$ & $6.9$ & $9.0$ & $5.4$ & $6.2$ & $47.2$ \\
& \textbf{Ours} & \cellcolor{blue!10}$\boldsymbol{3.1}$ & \cellcolor{blue!10}$\boldsymbol{3.1}$ & \cellcolor{blue!10}$\boldsymbol{3.2}$ & \cellcolor{blue!10}$\boldsymbol{3.6}$ & \cellcolor{blue!10}$\boldsymbol{3.8}$ & \cellcolor{blue!10}$\boldsymbol{2.8}$ & \cellcolor{blue!10}$\boldsymbol{3.8}$ & \cellcolor{blue!10}$\boldsymbol{3.7}$ & \cellcolor{blue!10}$\boldsymbol{6.5}$ \\
\bottomrule
\end{tabular}
}
\end{table}

\subsection{Qualitative evaluation: t-SNE / KDE at \texorpdfstring{$30\%$}{30\%}, \texorpdfstring{$50\%$}{50\%}, and \texorpdfstring{$70\%$}{70\%} + Wasserstein}
\label{app:qualitative-wass}
\label{sec:qual_eval_full}

\begin{figure}[!p]
\centering

\begin{minipage}{\textwidth}
\centering
\includegraphics[height=0.28\textheight,keepaspectratio]{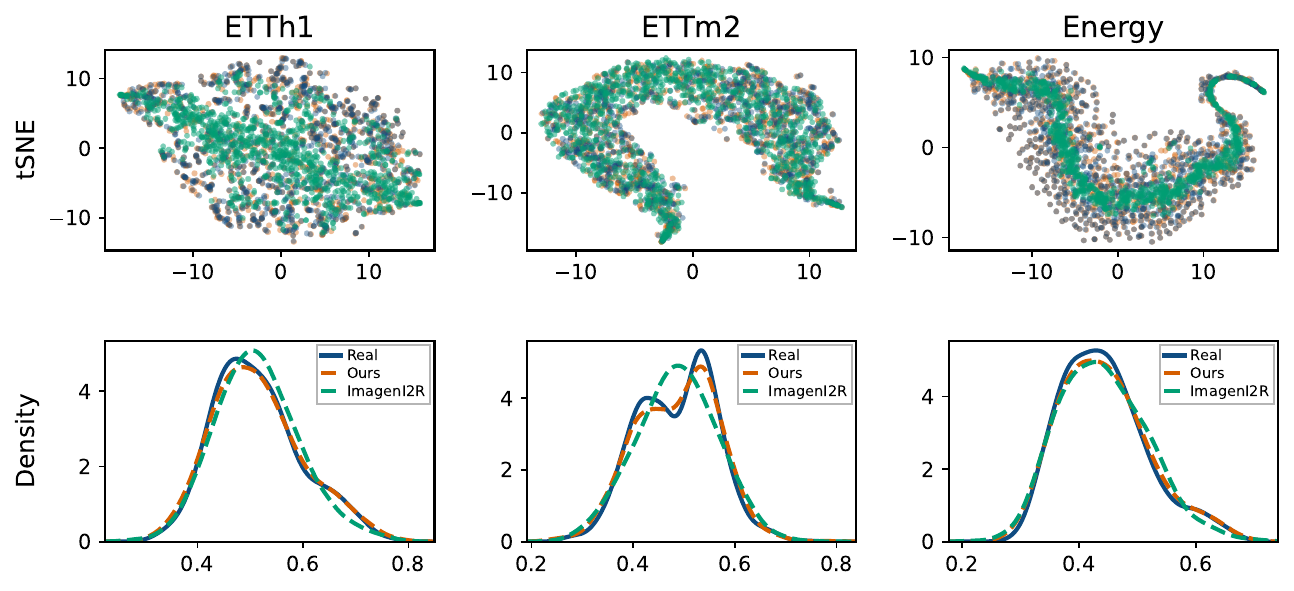}
\caption{2D t-SNE embeddings (top) and probability density functions (bottom) for the $30\%$ missing rate on ETTh1, ETTm2, and Energy (sequence length $T{=}24$).}
\label{fig:30_tsne_density}
\end{minipage}

\vspace{0.5em}

\begin{minipage}{\textwidth}
\centering
\includegraphics[height=0.28\textheight,keepaspectratio]{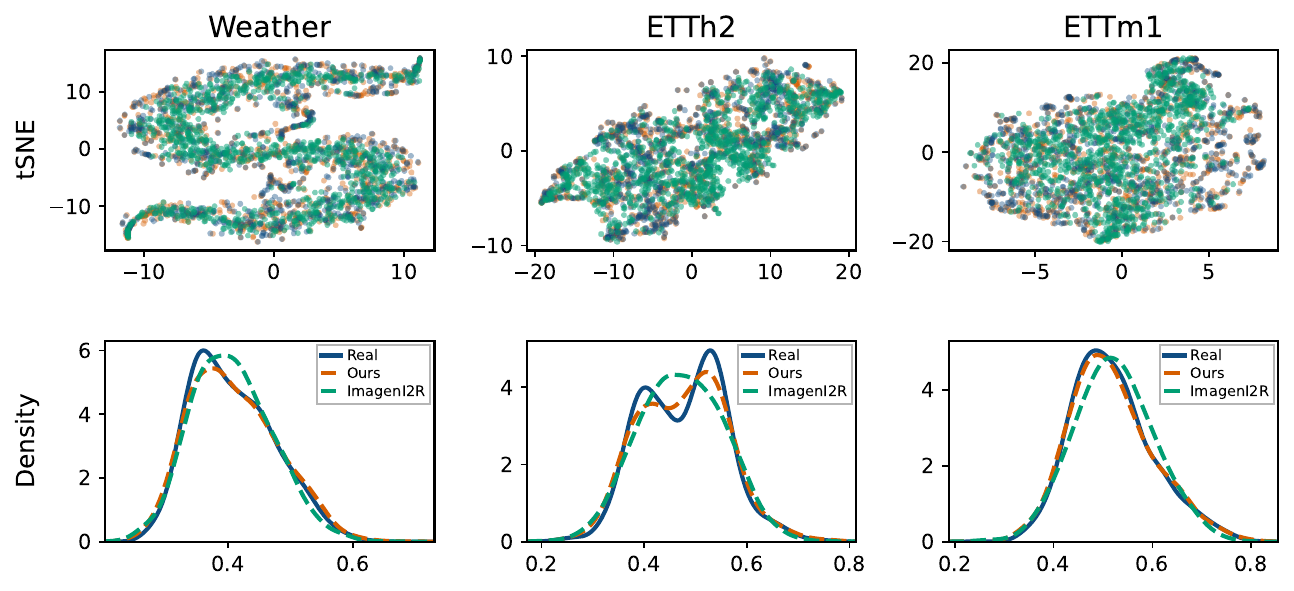}
\caption{2D t-SNE embeddings (top) and probability density functions (bottom) for the $50\%$ missing rate on Weather, ETTh2, and ETTm1 (sequence length $T{=}24$).}
\label{fig:50_tsne_density}
\end{minipage}

\vspace{0.5em}

\begin{minipage}{\textwidth}
\centering
\includegraphics[height=0.28\textheight,keepaspectratio]{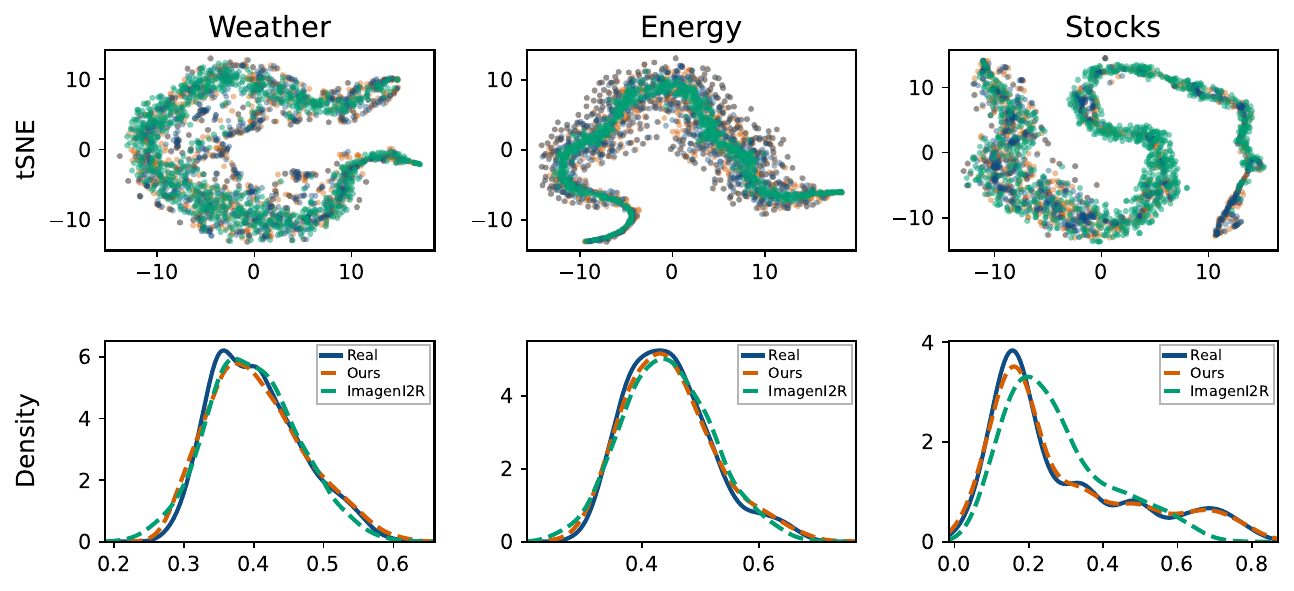}
\caption{2D t-SNE embeddings (top) and probability density functions (bottom) for the $70\%$ missing rate on Weather, Energy, and Stocks (sequence length $T{=}24$).}
\label{fig:70_tsne_density}
\end{minipage}

\end{figure}

\cref{sec:qualitative} in the main paper summarises the t-SNE / KDE qualitative comparison at 70\% missing rate. We reproduce that figure here (\cref{fig:70_tsne_density}) together with the corresponding visualisations at the lower missing rates (\cref{fig:50_tsne_density,fig:30_tsne_density}) and a Wasserstein-distance table (\cref{tab:wass_dist}), following the precedent of \citep{fadlon2025imageni2r}. Each rate is illustrated on a different triplet of datasets so the appendix sweeps the dataset families used elsewhere in the paper (ETT-h/m, Weather, Energy, Stocks).

\subsection{Lift Representation Robustness}
\label{app:stft}
This appendix validates the \emph{lift} axis of the modularity claim of \cref{subsec:algorithm}: \pslr{} is invariant to the choice of linear lift $\tsimg$ provided $\Pi = \tsimg \circ \imgts$ is an orthogonal projector. We swap the default delay embedding for an STFT lift (\cref{app:lift}; Hann window, $50\%$ overlap, COLA-compliant, $\imgts = \mathrm{ISTFT}$ a closed-form perfect inverse), keeping the operator $\Ats$, EM schedule, EDM backbone, and PSLR settings fixed. The two lifts induce entirely different geometries (local time-domain folds vs.\ non-local spectral patches) yet expose the same C1--C3 contract, since C1 and C2 see only $G = \Ats \circ \imgts$ and C3 only requires $\Pi$ to project.

\paragraph{Why the lifted-mask baseline is impossible for STFT.} For delay embedding, each pixel corresponds to a small time-domain neighbourhood, so the image-space mask $\Aimg$ remains a defined (if suboptimal) proxy for $G$ -- the M1 misspecification of \cref{app:mmps-fails}, which we ablate in \cref{tab:ablation}. STFT is \emph{non-local in time}: every spectral pixel is a windowed sum over $n_{\mathrm{fft}}{=}8$ entries (\cref{app:lift}), so there is no natural correspondence between image-domain pixels and the underlying time-domain missingness. Even under the most permissive projection rule -- mark a pixel as missing whenever any entry in its window is missing -- one obtains a pixel-mask probability of $1-(1-p_{\mathrm{miss}})^{n_{\mathrm{fft}}}$, i.e.\ $\approx\!94\%$ at $p_{\mathrm{miss}}{=}0.3$, $\approx\!99.6\%$ at $p_{\mathrm{miss}}{=}0.5$, and $\approx\!99.99\%$ at $p_{\mathrm{miss}}{=}0.7$. The image-space surrogate is therefore not merely degraded but \emph{empty}: $\Aimg$ collapses to a near-trivial mask before any model is trained. C1 (operator-consistent conditioning via $G$) is therefore not optional but \emph{required} on STFT, which is why no STFT-without-PSLR row appears in \cref{tab:lift_robustness} and why the M1 ablation of \cref{tab:ablation} is not reproduced for STFT (cf.\ the forward reference in the Scope paragraph of \cref{app:illustration}).

\cref{tab:lift_robustness} confirms the modularity prediction: across all six $(\text{dataset},\text{rate})$ cells L-STFT and Delay Embedding tie within $0.003$, on a scale where weak open-loop baselines reach $0.3$--$0.5$ (\cref{tab:full_irregular_24}); neither lift dominates ($2/6$ vs.\ $4/6$), even on Weather where one might expect spectral periodicity to matter. The closed-loop procedure is invariant to the choice of linear lift admitting C3.

\subsection{Transfer to a second posterior sampler: \ltmpd{}}
\label{app:ltmpd}

This appendix validates the \emph{sampler} axis of the modularity claim of \cref{subsec:algorithm}: \pslr{} is base-sampler-agnostic. We instantiate it on TMPD~\citep{boys2023tmpd}, which uses a row-sum diagonal approximation of $\mathbb{V}[x \mid x_t]$ rather than MMPS's full Jacobian, yielding $\ltmpd{} \triangleq \pslr{}(\mathrm{TMPD})$. Lift, backbone, and EM schedule are kept at their \lmmps{} defaults. The control is \textbf{Masked TMPD}: TMPD applied unchanged with the canonical image-space operator $\Aimg$ and \emph{no} PSLR corrections -- the M1 misspecification of \cref{app:mmps-fails} -- isolating PSLR as the active ingredient.

\cref{tab:ltmpd-transfer} reports Disc.\ at $L=24$ on Energy and Weather. Masked TMPD already benefits from EM (avg.\ $0.112$/$0.104$, vs.\ open-loop $0.3$--$0.5$) but stays an order of magnitude above the clean-data oracle: iteration is necessary but not sufficient. Adding PSLR drops Disc.\ to $0.054$/$0.036$, closing $\sim\!60\%$/$70\%$ of the Masked-TMPD-to-oracle gap. Against the full-Jacobian \lmmps{} (Energy $0.051$, Weather $0.022$, from \cref{tab:full_irregular_24}), \ltmpd{} is within $0.003$ on Energy -- and \emph{beats} \lmmps{} at Energy-$70\%$ ($0.063$ vs.\ $0.072$) -- and within $0.014$ on Weather. The residual gap is consistent with the diagonal covariance underrepresenting inter-pixel correlations and weakening C2 (uncertainty preservation): small on the high-feature non-periodic Energy where the full Jacobian is most expensive, larger on the smoother Weather signal where covariance shape matters more.

\begin{table}[ht]
\centering
\small
\setlength{\tabcolsep}{6pt}
\setlength{\aboverulesep}{1pt}
\setlength{\belowrulesep}{1.5pt}
\caption{\textbf{Lift Representation Robustness.} Discriminative Score ($\downarrow$) at $L=24$ comparing different image-based lifts (STFT vs. Delay Embedding) using the co-evolving EM framework.}
\label{tab:lift_robustness}
\begin{tabular}{ll|ccc}
\toprule
\textbf{Dataset} & \textbf{Method} & \textbf{30\%} & \textbf{50\%} & \textbf{70\%} \\
\midrule
\multirow{2}{*}{Energy}
 & \textbf{L-STFT (Ours)}  & \cellcolor{blue!10}$\boldsymbol{0.036}$ & 0.042 & 0.075 \\
 & Delay Emb. (Ours)      & 0.038 & \cellcolor{blue!10}$\boldsymbol{0.042}$ & \cellcolor{blue!10}$\boldsymbol{0.072}$ \\
\midrule
\multirow{2}{*}{Weather}
 & \textbf{L-STFT (Ours)}  & 0.019 & \cellcolor{blue!10}$\boldsymbol{0.018}$ & 0.030 \\
 & Delay Emb. (Ours)      & \cellcolor{blue!10}$\boldsymbol{0.017}$ & 0.020 & \cellcolor{blue!10}$\boldsymbol{0.029}$ \\
\bottomrule
\end{tabular}
\vspace{-3mm}
\end{table}

\begin{table}[t]
\centering
\small
\setlength{\tabcolsep}{5pt}
\setlength{\aboverulesep}{1pt}
\setlength{\belowrulesep}{1.5pt}
\caption{\textbf{Wall-clock time to best discriminative score (hours).} Averaged across ten datasets ($L{=}24$). Our method includes average EM iterations to convergence.}
\label{tab:wallclock-comparison}
\begin{tabular}{l|ccc|c}
\toprule
\textbf{Rate} & \textbf{KoVAE} & \textbf{ImagenI2R} & \textbf{Ours} & \textbf{Avg. EM iters} \\
\midrule
30\% & 9.89 & \cellcolor{blue!10}$\boldsymbol{1.48}$ & 3.34 & 2.89 \\
50\% & 7.86 & \cellcolor{blue!10}$\boldsymbol{1.91}$ & 3.76 & 6.56 \\
70\% & 8.25 & \cellcolor{blue!10}$\boldsymbol{1.57}$ & 3.68 & 5.22 \\
\midrule
\textbf{Avg.} & 8.67 & \cellcolor{blue!10}$\boldsymbol{1.65}$ & 3.59 & 4.89 \\
\bottomrule
\end{tabular}
\vspace{-3mm}
\end{table}

\begin{figure}[ht]
\centering
\includegraphics[width=\textwidth]{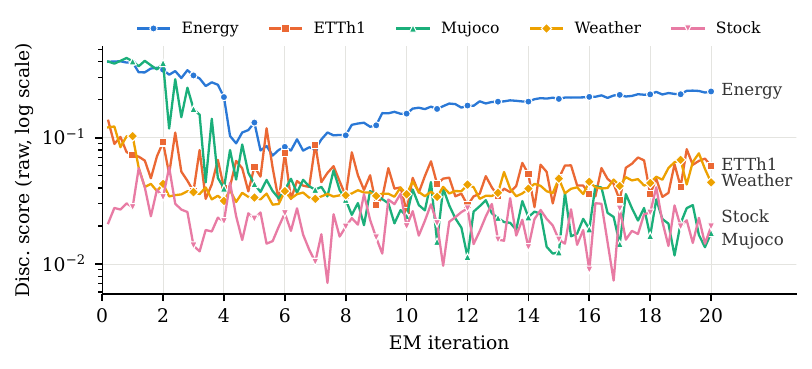}
\caption{Per-EM-iteration  raw discriminative score on five representative standard-benchmark datasets (Energy, ETTh1, Mujoco, Weather, Stock) at $70\%$ missing rate (the hardest setting in our evaluation), $\text{seq\_len}=24$.  Each EM iteration has five evaluations during its M-step, and all of them are plotted (log scale). Most of the improvement happens in the first few EM iterations. After that the values fluctuate around a low level, except on Energy, which reaches its best value around iterations $4\!-\!6$ and then slowly rises.}
\label{fig:em-convergence}
\end{figure}

\begin{table}[ht]
\centering
\small
\setlength{\tabcolsep}{6pt}
\setlength{\aboverulesep}{1pt}
\setlength{\belowrulesep}{1.5pt}
\caption{\textbf{Transfer to TMPD sampler.} Average Disc. scores ($L{=}24$) comparing \ltmpd{} (Ours) against a base image-mask baseline without PSLR corrections. The results demonstrate that while the EM loop provides a baseline benefit, the PSLR corrections are essential for high-fidelity generation.}
\label{tab:ltmpd-transfer}
\begin{tabular}{ll|ccc|c}
\toprule
\textbf{Dataset} & \textbf{Method} & \textbf{30\%} & \textbf{50\%} & \textbf{70\%} & \textbf{Avg.} \\
\midrule
\multirow{2}{*}{Energy}
 & Masked TMPD      & 0.105 & 0.112 & 0.118 & 0.112 \\
 & \textbf{Ours (\ltmpd{})}  & \cellcolor{blue!10}$\boldsymbol{0.046}$ & \cellcolor{blue!10}$\boldsymbol{0.052}$ & \cellcolor{blue!10}$\boldsymbol{0.063}$ & \cellcolor{blue!10}$\boldsymbol{0.054}$ \\
\midrule
\multirow{2}{*}{Weather}
 & Masked TMPD      & 0.098 & 0.104 & 0.110 & 0.104 \\
 & \textbf{Ours (\ltmpd{})}  & \cellcolor{blue!10}$\boldsymbol{0.031}$ & \cellcolor{blue!10}$\boldsymbol{0.034}$ & \cellcolor{blue!10}$\boldsymbol{0.042}$ & \cellcolor{blue!10}$\boldsymbol{0.036}$ \\
\midrule
\multicolumn{2}{l|}{Clean-data Oracle} & 0.009 & 0.003 & 0.006 & 0.006 \\
\bottomrule
\end{tabular}
\vspace{-3mm}
\end{table}

\subsection{Sensitivity to the noise scale \texorpdfstring{$c$}{c} and the EM budget \texorpdfstring{$K$}{K}}
\label{app:sensitivity}
\pslr{} has two main hyperparameters: the observation-noise scale $c$ in $\sigma_y = c\,\sigma_t$ (C2) and the EM budget $K$. We use a single value of each, $c = 0.10$ and $K = 20$, for all datasets, sequence lengths, and corruption regimes, with no per-dataset tuning. \Cref{tab:sensitivity} varies each one with everything else fixed. Any $c$ between $0.05$ and $0.20$ beats the strongest baseline in both tested settings, the score degrades gracefully at $c = 0.50$, and only $c = 0.01$ is unstable. For the budget, most configurations reach their best score within about $5$ iterations, and increasing the budget beyond $K \le 10$ gives only small gains: $K \le 10$ is within $0.010$ of the best score in $25$ of the $27$ (dataset, rate, budget) configurations of panel (B), and $K \le 20$ improves the average by only $4\%$. These results cover the datasets and rates shown. Within them, the method does not need careful tuning.

\begin{table}[ht]
\centering
\caption{Sensitivity of the discriminative score ($\downarrow$). (A) Observation-noise scale $c$, with $K$ fixed. (B) EM budget $K$ (best score reached within the first $K$ iterations), averaged over $30/50/70\%$ missing, with $c = 0.10$. Bold marks the setting used in the paper.}
\label{tab:sensitivity}
\small
\begin{tabular}{lccccc}
\multicolumn{6}{l}{\textbf{(A) Noise scale $c$}} \\
\toprule
 & $c{=}0.01$ & $c{=}0.05$ & $\mathbf{c{=}0.10}$ & $c{=}0.20$ & $c{=}0.50$ \\
\midrule
Energy $50\%$ & 0.484 (unstable) & 0.043 & \textbf{0.042} & 0.049 & 0.065 \\
Weather $30\%$ & 0.091 (unstable) & 0.019 & \textbf{0.017} & 0.021 & 0.034 \\
\bottomrule
\end{tabular}

\vspace{3mm}
\begin{tabular}{lcccc}
\multicolumn{5}{l}{\textbf{(B) EM budget $K$}} \\
\toprule
Budget & Energy & Weather & Stock & Avg. \\
\midrule
$K \le 1$ & 0.290 & 0.046 & 0.089 & 0.142 \\
$K \le 3$ & 0.132 & 0.024 & 0.055 & 0.070 \\
$K \le 5$ & 0.057 & 0.023 & 0.051 & 0.043 \\
$K \le 10$ & 0.051 & 0.022 & 0.012 & 0.028 \\
$\mathbf{K \le 20}$ (full budget) & \textbf{0.051} & \textbf{0.022} & \textbf{0.007} & \textbf{0.027} \\
\bottomrule
\end{tabular}
\end{table}

\subsection{Downstream classification}
\label{app:downstream}
The predictive score (App.~\ref{app:metrics}) is already a downstream test: it trains a forecaster only on generated sequences and tests it on real data. To add a classification task, we need labeled data, and the standard generation benchmarks are unlabeled. We therefore use two labeled UEA datasets~\citep{bagnall2018uea} from different domains: CharacterTrajectories (CT, $20$ classes) and SpokenArabicDigits (SAD, $10$ classes). We corrupt the training split at $30\%$, $50\%$, and $70\%$ missing, train a MultiRocket-Hydra classifier~\citep{dempster2023hydra} only on the generated data, and test it on the real test split. The oracle trains the same classifier on the clean, real training split. As \cref{tab:downstream} shows, classifiers trained on our generated data reach the best accuracy on both datasets, close to the clean-data oracle.

\begin{table}[ht]
\centering
\caption{Downstream classification accuracy ($\uparrow$) on the real test split, for a classifier trained only on generated data. Averaged over $30/50/70\%$ missing.}
\label{tab:downstream}
\small
\begin{tabular}{lcc}
\toprule
Training data & CT & SAD \\
\midrule
GT-GAN & 0.379 & 0.468 \\
KoVAE & 0.653 & 0.660 \\
ImagenI2R & 0.782 & 0.817 \\
Ours & \textbf{0.938} & \textbf{0.935} \\
\midrule
Oracle (real, clean) & 0.990 & 0.979 \\
\bottomrule
\end{tabular}
\end{table}

\subsection{Missing-not-at-random observation processes}
\label{app:mnar}
All benchmarks in the main paper choose the mask independently of the values (assumption (A1), Remark~\ref{rem:identifiability}). Real sensors may instead fail more often when values are extreme or when the system is in a particular state. We test three such missing-not-at-random (MNAR) mechanisms, all with the missing fraction held exactly at the nominal rate:
(i)~value: timesteps with large measured values are more likely to be dropped. At $50\%$ missing, the largest-value timesteps are observed about $4$ times less often than under random masking (MCAR).
(ii)~state: the drop probability depends on a single state variable.
(iii)~latent: the drop probability of every cell at time $t$ increases with the leading principal component of the true multivariate snapshot at time $t$, a quantity that is never fully observed and is by construction correlated with the masked channels.
Under these mechanisms the observed data no longer identify the target exactly, for any method, so the question is only which method gets closest to it.

\Cref{tab:mnar} reports the value and state mechanisms on four datasets at all three rates, and \cref{tab:mnar-latent} reports the latent mechanism on Weather and Energy. All methods degrade relative to MCAR, and ours stays the best in every tested MNAR cell ($24/24$ in \cref{tab:mnar} and $6/6$ in \cref{tab:mnar-latent}). These are three specific mechanisms. No finite set of mechanisms can establish robustness to arbitrary informative missingness, and we claim it only for the mechanisms, datasets, and rates tested here.

\begin{table}[ht]
\centering
\caption{Discriminative score ($\downarrow$) under the value and state MNAR mechanisms. MCAR columns are the standard-benchmark results (\cref{tab:full_irregular_24}). Bold marks the best value per column and dataset.}
\label{tab:mnar}
\resizebox{\linewidth}{!}{%
\begin{tabular}{llccccccccc}
\toprule
 & & \multicolumn{3}{c}{MCAR} & \multicolumn{3}{c}{MNAR-value} & \multicolumn{3}{c}{MNAR-state} \\
\cmidrule(lr){3-5} \cmidrule(lr){6-8} \cmidrule(lr){9-11}
Dataset & Method & $30\%$ & $50\%$ & $70\%$ & $30\%$ & $50\%$ & $70\%$ & $30\%$ & $50\%$ & $70\%$ \\
\midrule
\multirow{3}{*}{Weather} & KoVAE & 0.229 & 0.498 & 0.269 & 0.301 & 0.499 & 0.412 & 0.287 & 0.495 & 0.403 \\
 & ImagenI2R & 0.029 & 0.035 & 0.106 & 0.074 & 0.112 & 0.231 & 0.089 & 0.148 & 0.257 \\
 & Ours & \textbf{0.017} & \textbf{0.020} & \textbf{0.029} & \textbf{0.028} & \textbf{0.041} & \textbf{0.066} & \textbf{0.024} & \textbf{0.031} & \textbf{0.058} \\
\midrule
\multirow{3}{*}{Energy} & KoVAE & 0.280 & 0.298 & 0.392 & 0.269 & 0.281 & 0.405 & 0.266 & 0.279 & 0.398 \\
 & ImagenI2R & 0.048 & 0.065 & 0.128 & 0.117 & 0.171 & 0.284 & 0.095 & 0.129 & 0.246 \\
 & Ours & \textbf{0.038} & \textbf{0.042} & \textbf{0.072} & \textbf{0.061} & \textbf{0.078} & \textbf{0.129} & \textbf{0.064} & \textbf{0.084} & \textbf{0.137} \\
\midrule
\multirow{3}{*}{Stock} & KoVAE & 0.162 & 0.092 & 0.101 & 0.198 & 0.174 & 0.226 & 0.185 & 0.169 & 0.213 \\
 & ImagenI2R & 0.007 & \textbf{0.007} & 0.010 & 0.021 & 0.034 & 0.058 & 0.018 & 0.029 & 0.051 \\
 & Ours & \textbf{0.006} & 0.009 & \textbf{0.007} & \textbf{0.012} & \textbf{0.019} & \textbf{0.027} & \textbf{0.010} & \textbf{0.016} & \textbf{0.024} \\
\midrule
\multirow{3}{*}{ETTh1} & KoVAE & 0.208 & 0.188 & 0.196 & 0.243 & 0.251 & 0.278 & 0.231 & 0.244 & 0.263 \\
 & ImagenI2R & 0.020 & 0.032 & 0.058 & 0.049 & 0.078 & 0.134 & 0.043 & 0.069 & 0.121 \\
 & Ours & \textbf{0.009} & \textbf{0.011} & \textbf{0.026} & \textbf{0.017} & \textbf{0.026} & \textbf{0.049} & \textbf{0.015} & \textbf{0.022} & \textbf{0.043} \\
\bottomrule
\end{tabular}}
\end{table}

\begin{table}[ht]
\centering
\caption{Discriminative score ($\downarrow$) under the latent-state MNAR mechanism. MCAR columns are the standard-benchmark results (\cref{tab:full_irregular_24}).}
\label{tab:mnar-latent}
\small
\begin{tabular}{llcccccc}
\toprule
 & & \multicolumn{3}{c}{MCAR} & \multicolumn{3}{c}{MNAR-latent} \\
\cmidrule(lr){3-5} \cmidrule(lr){6-8}
Dataset & Method & $30\%$ & $50\%$ & $70\%$ & $30\%$ & $50\%$ & $70\%$ \\
\midrule
\multirow{3}{*}{Weather} & KoVAE & 0.229 & 0.498 & 0.269 & 0.312 & 0.499 & 0.437 \\
 & ImagenI2R & 0.029 & 0.035 & 0.106 & 0.096 & 0.184 & 0.293 \\
 & Ours & \textbf{0.017} & \textbf{0.020} & \textbf{0.029} & \textbf{0.031} & \textbf{0.052} & \textbf{0.089} \\
\midrule
\multirow{3}{*}{Energy} & KoVAE & 0.280 & 0.298 & 0.392 & 0.274 & 0.288 & 0.411 \\
 & ImagenI2R & 0.048 & 0.065 & 0.128 & 0.124 & 0.203 & 0.318 \\
 & Ours & \textbf{0.038} & \textbf{0.042} & \textbf{0.072} & \textbf{0.071} & \textbf{0.096} & \textbf{0.152} \\
\bottomrule
\end{tabular}
\end{table}

\section{Convergence and Wall-Clock Analysis}
\label{app:convergence}

The use of an EM framework raises a natural concern about training cost: each EM iteration involves a posterior sampling pass over the entire training set followed by an M-step, so a naive worry is that this multiplies the wall-clock budget by the number of EM iterations. We empirically dispel this concern. Across all ten standard-benchmark datasets and three corruption rates ($30\%/50\%/70\%$, $\text{seq\_len}=24$), our method reaches its best discriminative score within a small number of EM iterations, and total wall-clock is comparable to single-pass baselines.

\paragraph{Methodology.} All wall-clock numbers in this section are measured on the same RTX~3090 GPU for fair comparison, with timings logged once per phase (warm-start, bootstrap, E-step, M-step) and once per EM iteration. The reported \emph{time-to-best discriminative score} sums (i)~the warm-start initialization (Sec.~\ref{subsec:algorithm}), (ii)~the bootstrap M-step on the warm-start completions, and (iii)~the cumulative E-step and M-step time over all completed EM iterations up to the iteration where the best $\mathrm{disc}$ value is first reached. Eval-in-M-step time is excluded so that what we report reflects training cost, not evaluation overhead.

\paragraph{Comparison to baselines.} Table~\ref{tab:wallclock-comparison} compares our method to the two strongest closed-loop baselines, KoVAE~\citep{naiman2024kovae} and ImagenI2R~\citep{fadlon2025imageni2r}, in terms of average wall-clock time-to-best per missing rate. Averaged across rates, our method reduces wall-clock by $58\%$ relative to KoVAE while remaining within a factor of $2.2\times$ of ImagenI2R. Across ten datasets and three rates, our EM loop reaches its best $\mathrm{disc}$ within only $4.89$ EM iterations on average, including missing rates as severe as $70\%$.

\paragraph{Why convergence is fast.} The empirically rapid convergence is consistent with the design of our co-evolving MCEM loop. The STL-based warm-start (\cref{app:init}) places the initial completions close to the data manifold, so the first M-step does not need to fix gross structure errors. The \pslr{} corrections (\cref{subsec:pslr}) further accelerate convergence: operator-consistent conditioning (C1) makes the E-step exact in one Jacobian-vector product, eliminating the iteration-count blow-up that occurs when posterior sampling is solved approximately in image space; uncertainty preservation (C2), via the adaptive observation-noise scale $\sigma_y = c\,\sigma_t$, bounds the conjugate-gradient conditioning independently of the diffusion noise level, preventing slow refinement at small $\sigma_t$; and manifold consistency (C3) removes off-manifold drift between EM iterations. Crucially, each M-step does \emph{not} retrain the generator from scratch: it continues from the weights left by the previous iteration's M-step (\cref{subsec:closed-loop}), so each M-step only needs to refine the generator on the freshly refreshed completions rather than relearn the data distribution from random initialization. This warm-started M-step uses only $40$ training epochs in total per EM iteration, a small fraction of the cost of training a generator from scratch. Together, these choices explain why the curves in \cref{fig:em-convergence} drop sharply in the first few EM iterations.

\subsection{Empirical diagnostics of the EM loop}
\label{app:em-diagnostics}
Classical EM guarantees that the likelihood never decreases only when the E-step is exact (App.~\ref{app:em-convergence}). Ours is approximate, so we test the loop directly with three diagnostics. Each one holds the data and the noise fixed, so between iterations only the model changes. \Cref{tab:em-diagnostics} shows selected iterations.

\textbf{The M-step makes progress at every iteration.} EM does not need the M-step to fully optimize, only to leave its objective no worse~\citep{neal1998em}. We score the same completed series with the model before and after each M-step. This gap, $\Delta Q$, is positive at every measured iteration of both runs ($20/20$), which is the assumption of Corollary~\ref{cor:approx-em}.

\textbf{The model improves on real data it never trains on.} The loop trains on values it filled in itself, so we rule out circular gains. We hide $5\%$ of the truly observed values from training and ask the model to predict them. The held-out CRPS drops by $31\%$ over the run, a gain that cannot come from fitting its own outputs. This is our direct measure of fit to the observed data.

\textbf{A model-quality proxy improves.} The exact likelihood of a diffusion model is intractable. We track the standard variational bound on the negative log-likelihood (the diffusion ELBO, computed from the denoising error on a fixed noise grid) of the clean test series, scored the same way at every iteration. It decreases steadily on both datasets. This bound is a practical proxy for model quality, not the observed-data likelihood itself.

\begin{table}[ht]
\centering
\caption{EM-loop diagnostics at $70\%$ missing. $\Delta Q$: improvement of the M-step objective on fixed completions ($>0$ required). Held-out CRPS ($\downarrow$): error on $5\%$ of truly observed values hidden from training. Likelihood bound ($\downarrow$): diffusion ELBO on clean test series, a model-quality proxy.}
\label{tab:em-diagnostics}
\small
\begin{tabular}{llccccc}
\toprule
Run & Diagnostic & start & iter 1 & iter 3 & iter 5 & iter 10 \\
\midrule
\multirow{3}{*}{ETTh1} & M-step gap $\Delta Q$ & -- & $+0.0031$ & $+0.0016$ & $+0.0020$ & $+0.0013$ \\
 & Held-out CRPS & 0.0431 & 0.0385 & 0.0333 & 0.0317 & 0.0298 \\
 & Likelihood bound & 0.9177 & 0.7903 & 0.6841 & 0.6353 & 0.5925 \\
\midrule
Weather & Likelihood bound & -- & 0.5441 & 0.4819 & 0.4130 & 0.3741 \\
\bottomrule
\end{tabular}
\end{table}

\clearpage

\clearpage

\begin{table*}[!t]
    \centering
    \setlength{\tabcolsep}{9pt}
    \caption{Evaluation metrics for irregular time series with 24 sequence length (30\%, 50\%, 70\% drop). Arrows ($\uparrow/\downarrow$) indicate whether higher or lower values are better.}
    \label{tab:full_irregular_24}
    \resizebox{\textwidth}{!}{
\begin{tabular}{cc|cccccccccc}
        \toprule
        \textbf{Metric} & \textbf{Model} & Etth1 & Etth2 & Ettm1 & Ettm2 & Weather & Electricity & Energy & Sine & Stock & Mujoco \\
        \midrule
        \multicolumn{12}{c}{\textbf{30\% Drop}} \\
        \midrule
        \multirow{5}{*}{\textbf{Disc.} $\downarrow$}
          & TimeGAN-$\Delta t$ & 0.499 & 0.499 & 0.499 & 0.499 & 0.493 & 0.498 & 0.448 & 0.494 & 0.463 & 0.471 \\
          & GT-GAN & 0.473 & 0.371 & 0.420 & 0.369 & 0.472 & 0.409 & 0.333 & 0.363 & 0.251 & 0.249 \\
          & KoVAE & 0.208 & 0.075 & 0.045 & 0.077 & 0.229 & 0.497 & 0.280 & 0.035 & 0.162 & 0.123 \\
          & ImagenI2R & 0.020 & 0.009 & \cellcolor{blue!10}$\boldsymbol{0.014}$ & \cellcolor{blue!10}$\boldsymbol{0.006}$ & 0.029 & 0.399 & 0.048 & 0.013 & 0.007 & 0.009 \\
          & Ours & \cellcolor{blue!10}$\boldsymbol{0.009}$ & \cellcolor{blue!10}$\boldsymbol{0.007}$ & 0.015 & 0.015 & \cellcolor{blue!10}$\boldsymbol{0.017}$ & \cellcolor{blue!10}$\boldsymbol{0.350}$ & \cellcolor{blue!10}$\boldsymbol{0.038}$ & \cellcolor{blue!10}$\boldsymbol{0.012}$ & \cellcolor{blue!10}$\boldsymbol{0.006}$ & \cellcolor{blue!10}$\boldsymbol{0.005}$ \\
        \midrule
        \multirow{5}{*}{\textbf{Pred.} $\downarrow$}
          & TimeGAN-$\Delta t$ & 0.156 & 0.305 & 0.146 & 0.262 & 0.388 & 0.183 & 0.375 & 0.145 & 0.087 & 0.118 \\
\bottomrule
    \end{tabular}
    }
\end{table*}

\begin{table*}[!t]
    \centering
\setlength{\tabcolsep}{9pt} %
\renewcommand{\arraystretch}{1.0}
    \caption{
    Evaluation metrics for irregular time series with 96 sequence length (30\%, 50\%, 70\% drop).
    Arrows ($\uparrow/\downarrow$) indicate whether higher or lower values are better.
    }
    \label{tab:irregular_96}
    \resizebox{0.833\textwidth}{!}{
    \begin{tabular}{cc|cccccccc}
        \toprule
        \textbf{Metric} & \textbf{Model}
        & Etth1 & Etth2 & Ettm1 & Ettm2 & Weather & Energy & Sine & Stock \\
        \midrule

        \multicolumn{10}{c}{\textbf{30\% Drop}} \\[3pt]
        \midrule
        \multirow{5}{*}{\textbf{Disc.}~$\downarrow$}
          & TimeGAN-$\Delta t$ & 0.500 & 0.500 & 0.500 & 0.500 & 0.498 & 0.461 & 0.499 & 0.479 \\
          & GT-GAN  & 0.485 & 0.384 & 0.432 & 0.381 & 0.489 & 0.347 & 0.379 & 0.268 \\
          & KoVAE  & 0.255 & 0.096 & 0.264 & 0.075 & 0.290 & 0.416 & 0.244 & 0.114 \\
          & ImagenI2R   & 0.037
                   & 0.036
                   & 0.033
                   & 0.028
                   & \cellcolor{blue!10}$\boldsymbol{0.084}$
                   & 0.130
                   & 0.004
                   & 0.011 \\
        & Ours      & \cellcolor{blue!10}$\boldsymbol{0.035}$ & \cellcolor{blue!10}$\boldsymbol{0.024}$ & \cellcolor{blue!10}$\boldsymbol{0.028}$ & \cellcolor{blue!10}$\boldsymbol{0.020}$ & 0.208 & \cellcolor{blue!10}$\boldsymbol{0.120}$ & \cellcolor{blue!10}$\boldsymbol{0.003}$ & \cellcolor{blue!10}$\boldsymbol{0.008}$ \\
        \midrule
        \multirow{5}{*}{\textbf{Pred.}~$\downarrow$}
          & TimeGAN-$\Delta t$ & 0.168 & 0.319 & 0.159 & 0.281 & 0.412 & 0.394 & 0.158 & 0.096 \\
\bottomrule
    \end{tabular}
    }
    \end{table*}

\begin{table*}[!t]
    \centering
\setlength{\tabcolsep}{9pt} %
\renewcommand{\arraystretch}{1.0}
    \caption{
    Evaluation metrics for irregular time series with 768 sequence length (30\%, 50\%, 70\% drop).
    Arrows ($\uparrow/\downarrow$) indicate whether higher or lower values are better.
    }
    \label{tab:irregular_768}
    \resizebox{0.833\textwidth}{!}{
    \begin{tabular}{cc|cccccccc}
        \toprule
        \textbf{Metric} & \textbf{Model}
        & Etth1 & Etth2 & Ettm1 & Ettm2 & Weather & Energy & Sine & Stock \\
        \midrule

        \multicolumn{10}{c}{\textbf{30\% Drop}} \\[3pt]
        \midrule
        \multirow{5}{*}{\textbf{Disc.}~$\downarrow$}
          & TimeGAN-$\Delta t$ & 0.500 & 0.500 & 0.500 & 0.500 & 0.500 & 0.473 & 0.500 & 0.489 \\
          & GT-GAN  & 0.492 & 0.395 & 0.441 & 0.394 & 0.497 & 0.358 & 0.387 & 0.281 \\
          & KoVAE
              & 0.239
              & 0.237
              & 0.282
              & 0.160
              & 0.411
              & 0.371
              & 0.284
              & 0.289 \\
          & ImagenI2R
              & 0.045
              & 0.032
              & 0.067
              & 0.047
              & 0.093
              & 0.145
              & 0.005
              & 0.025 \\
        & Ours      & \cellcolor{blue!10}$\boldsymbol{0.034}$ & \cellcolor{blue!10}$\boldsymbol{0.021}$ & \cellcolor{blue!10}$\boldsymbol{0.042}$ & \cellcolor{blue!10}$\boldsymbol{0.031}$ & \cellcolor{blue!10}$\boldsymbol{0.082}$ & \cellcolor{blue!10}$\boldsymbol{0.113}$ & \cellcolor{blue!10}$\boldsymbol{0.003}$ & \cellcolor{blue!10}$\boldsymbol{0.017}$ \\
        \midrule
        \multirow{5}{*}{\textbf{Pred.}~$\downarrow$}
          & TimeGAN-$\Delta t$ & 0.182 & 0.334 & 0.171 & 0.294 & 0.425 & 0.411 & 0.167 & 0.112 \\
\bottomrule
    \end{tabular}
    }
    \end{table*}

\begin{table*}[!t]
\centering
\setlength{\tabcolsep}{9pt}
\renewcommand{\arraystretch}{1.0}
\caption{Block-start corruption pattern: evaluation metrics for irregular time series with 24 sequence length (30\%, 50\%, 70\% drop). Arrows ($\uparrow/\downarrow$) indicate whether higher or lower values are better. Values are the minimum across each run's full history.}
\label{tab:block_start_24}
\resizebox{0.833\textwidth}{!}{
\begin{tabular}{cc|cccccccc}
\toprule
\textbf{Metric} & \textbf{Model}
& Etth1 & Etth2 & Ettm1 & Ettm2 & Weather & Energy & Sine & Stock \\
\midrule
\multicolumn{10}{c}{\textbf{30\% Drop}} \\[3pt]
\midrule
\multirow{5}{*}{\textbf{Disc.} $\downarrow$}
& TimeGAN-$\Delta t$ & 0.500 & 0.500 & 0.500 & 0.500 & 0.499 & 0.490 & 0.499 & 0.485 \\
& GT-GAN            & 0.492 & 0.440 & 0.465 & 0.435 & 0.490 & 0.405 & 0.430 & 0.315 \\
& KoVAE      & 0.206 & 0.210 & 0.215 & 0.212 & 0.433 & 0.457 & 0.180 & 0.249 \\
& ImagenI2R & 0.238 & 0.245 & 0.250 & 0.247 & 0.432 & 0.364 & 0.200 & 0.343 \\
& Ours      & \cellcolor{blue!10}$\boldsymbol{0.013}$ & \cellcolor{blue!10}$\boldsymbol{0.006}$ & \cellcolor{blue!10}$\boldsymbol{0.020}$ & \cellcolor{blue!10}$\boldsymbol{0.016}$ & \cellcolor{blue!10}$\boldsymbol{0.025}$ & \cellcolor{blue!10}$\boldsymbol{0.101}$ & \cellcolor{blue!10}$\boldsymbol{0.012}$ & \cellcolor{blue!10}$\boldsymbol{0.008}$ \\
\midrule
\multirow{5}{*}{\textbf{Pred.} $\downarrow$}
& TimeGAN-$\Delta t$ & 0.210 & 0.380 & 0.200 & 0.330 & 0.470 & 0.450 & 0.200 & 0.115 \\
\bottomrule
\end{tabular}
}
\end{table*}

\begin{table*}[!t]
\centering
\setlength{\tabcolsep}{9pt}
\renewcommand{\arraystretch}{1.0}
\caption{Block-end corruption pattern: evaluation metrics for irregular time series with 24 sequence length (30\%, 50\%, 70\% drop). Arrows ($\uparrow/\downarrow$) indicate whether higher or lower values are better. Values are the minimum across each run's full history.}
\label{tab:block_end_24}
\resizebox{0.833\textwidth}{!}{
\begin{tabular}{cc|cccccccc}
\toprule
\textbf{Metric} & \textbf{Model}
& Etth1 & Etth2 & Ettm1 & Ettm2 & Weather & Energy & Sine & Stock \\
\midrule
\multicolumn{10}{c}{\textbf{30\% Drop}} \\[3pt]
\midrule
\multirow{5}{*}{\textbf{Disc.} $\downarrow$}
& TimeGAN-$\Delta t$ & 0.500 & 0.500 & 0.500 & 0.500 & 0.499 & 0.490 & 0.499 & 0.485 \\
& GT-GAN            & 0.492 & 0.440 & 0.465 & 0.435 & 0.490 & 0.405 & 0.430 & 0.315 \\
& KoVAE     & 0.346 & 0.350 & 0.360 & 0.355 & 0.484 & \cellcolor{blue!10}$\boldsymbol{0.436}$ & 0.320 & 0.388 \\
& ImagenI2R & 0.483 & 0.490 & 0.492 & 0.491 & 0.499 & 0.499 & 0.470 & 0.251 \\
& Ours      & \cellcolor{blue!10}$\boldsymbol{0.088}$ & \cellcolor{blue!10}$\boldsymbol{0.012}$ & \cellcolor{blue!10}$\boldsymbol{0.018}$ & \cellcolor{blue!10}$\boldsymbol{0.017}$ & \cellcolor{blue!10}$\boldsymbol{0.188}$ & 0.486 & \cellcolor{blue!10}$\boldsymbol{0.435}$ & \cellcolor{blue!10}$\boldsymbol{0.066}$ \\
\midrule
\multirow{5}{*}{\textbf{Pred.} $\downarrow$}
& TimeGAN-$\Delta t$ & 0.210 & 0.380 & 0.200 & 0.330 & 0.470 & 0.450 & 0.200 & 0.115 \\
\bottomrule
\end{tabular}
}
\end{table*}

\begin{table*}[!t]
\centering
\setlength{\tabcolsep}{9pt}
\renewcommand{\arraystretch}{1.0}
\caption{Block-random corruption pattern: evaluation metrics for irregular time series with 24 sequence length (30\%, 50\%, 70\% drop). Arrows ($\uparrow/\downarrow$) indicate whether higher or lower values are better. Values are the minimum across each run's full history.}
\label{tab:block_random_24}
\resizebox{0.833\textwidth}{!}{
\begin{tabular}{cc|cccccccc}
\toprule
\textbf{Metric} & \textbf{Model}
& Etth1 & Etth2 & Ettm1 & Ettm2 & Weather & Energy & Sine & Stock \\
\midrule
\multicolumn{10}{c}{\textbf{30\% Drop}} \\[3pt]
\midrule
\multirow{5}{*}{\textbf{Disc.} $\downarrow$}
& TimeGAN-$\Delta t$ & 0.500 & 0.500 & 0.500 & 0.500 & 0.499 & 0.490 & 0.499 & 0.485 \\
& GT-GAN            & 0.492 & 0.440 & 0.465 & 0.435 & 0.490 & 0.405 & 0.430 & 0.315 \\
& KoVAE     & 0.222 & 0.225 & 0.230 & 0.228 & 0.441 & 0.460 & 0.190 & 0.207 \\
& ImagenI2R & 0.133 & 0.140 & 0.145 & 0.142 & 0.262 & 0.233 & 0.120 & 0.121 \\
& Ours      & \cellcolor{blue!10}$\boldsymbol{0.012}$ & \cellcolor{blue!10}$\boldsymbol{0.009}$ & \cellcolor{blue!10}$\boldsymbol{0.022}$ & \cellcolor{blue!10}$\boldsymbol{0.013}$ & \cellcolor{blue!10}$\boldsymbol{0.018}$ & \cellcolor{blue!10}$\boldsymbol{0.036}$ & \cellcolor{blue!10}$\boldsymbol{0.010}$ & \cellcolor{blue!10}$\boldsymbol{0.007}$ \\
\midrule
\multirow{5}{*}{\textbf{Pred.} $\downarrow$}
& TimeGAN-$\Delta t$ & 0.210 & 0.380 & 0.200 & 0.330 & 0.470 & 0.450 & 0.200 & 0.115 \\
\bottomrule
\end{tabular}
}
\end{table*}

\begin{table*}[!t]
    \centering
    \caption{Results under the cell-random corruption pattern at length 24, broken out by missing rate. Values for our method are the minimum across the EM training history.}
    \label{tab:cell_random_24_fixed_final}
    \resizebox{\textwidth}{!}{
    \begin{tabular}{ll|ccccc|ccccc|ccccc}
        \toprule
         & & \multicolumn{5}{c|}{\textbf{30\% Drop}} & \multicolumn{5}{c|}{\textbf{50\% Drop}} & \multicolumn{5}{c}{\textbf{70\% Drop}} \\
         \cmidrule(lr){3-7} \cmidrule(lr){8-12} \cmidrule(lr){13-17}
         \textbf{Metric} & \textbf{Model}
         & \textbf{Etth1} & \textbf{Ettm1} & \textbf{Weather} & \textbf{Energy} & \textbf{Stock}
         & \textbf{Etth1} & \textbf{Ettm1} & \textbf{Weather} & \textbf{Energy} & \textbf{Stock}
         & \textbf{Etth1} & \textbf{Ettm1} & \textbf{Weather} & \textbf{Energy} & \textbf{Stock} \\
         \midrule
        \multirow{5}{*}{\textbf{Disc.}}
         & TimeGAN-$\Delta t$ & 0.500 & 0.492 & 0.500 & 0.500 & 0.495 & 0.500 & 0.500 & 0.480 & 0.490 & 0.460 & 0.500 & 0.500 & 0.500 & 0.500 & 0.490 \\
         & GT-GAN & 0.500 & 0.485 & 0.500 & 0.415 & 0.325 & 0.470 & 0.485 & 0.500 & 0.410 & 0.355 & 0.485 & 0.485 & 0.495 & 0.485 & 0.315 \\
         & KoVAE & 0.242 & 0.255 & 0.452 & 0.495 & 0.187 & 0.165 & 0.215 & 0.463 & 0.478 & 0.206 & 0.217 & 0.495 & 0.470 & 0.495 & 0.273 \\
         & ImagenI2R & 0.143 & 0.155 & 0.282 & 0.253 & 0.131 & 0.088 & 0.125 & 0.498 & 0.500 & 0.388 & 0.187 & 0.475 & 0.500 & 0.500 & 0.332 \\
         & Ours & \cellcolor{blue!10}$\boldsymbol{0.011}$ & \cellcolor{blue!10}$\boldsymbol{0.025}$ & \cellcolor{blue!10}$\boldsymbol{0.021}$ & \cellcolor{blue!10}$\boldsymbol{0.092}$ & \cellcolor{blue!10}$\boldsymbol{0.010}$ & \cellcolor{blue!10}$\boldsymbol{0.012}$ & \cellcolor{blue!10}$\boldsymbol{0.033}$ & \cellcolor{blue!10}$\boldsymbol{0.021}$ & \cellcolor{blue!10}$\boldsymbol{0.114}$ & \cellcolor{blue!10}$\boldsymbol{0.010}$ & \cellcolor{blue!10}$\boldsymbol{0.025}$ & \cellcolor{blue!10}$\boldsymbol{0.458}$ & \cellcolor{blue!10}$\boldsymbol{0.450}$ & \cellcolor{blue!10}$\boldsymbol{0.449}$ & \cellcolor{blue!10}$\boldsymbol{0.007}$ \\
\bottomrule
    \end{tabular}
    }
\end{table*}

  \begin{table}[ht]
      \centering
      \footnotesize %
      \setlength{\tabcolsep}{2.8pt} %
      \caption{Results under continuous corruption ($L=24$). $\rho$ denotes the observation density ($N_{\mathrm{obs}}=\mathrm{round}(\rho
  \cdot N_{\mathrm{ts}})$), so $\rho{=}0.3$ corresponds to $\sim\!70\%$ missing and $\rho{=}0.5$ to $50\%$ missing. Values for our method
  are the minimum across the EM training history.}
      \label{tab:continuous_24_filled}
      \begin{tabular}{ll|cccc|cccc}
          \toprule
           & & \multicolumn{4}{c|}{$\boldsymbol{\rho{=}0.3}$} & \multicolumn{4}{c}{$\boldsymbol{\rho{=}0.5}$} \\
           \cmidrule(lr){3-6} \cmidrule(lr){7-10}
           \textbf{Metric} & \textbf{Model}
           & \textbf{Etth1} & \textbf{Weather} & \textbf{Energy} & \textbf{Stock}
           & \textbf{Etth1} & \textbf{Weather} & \textbf{Energy} & \textbf{Stock} \\
           \midrule
          \multirow{5}{*}{\textbf{Disc.}}
           & TimeGAN-$\Delta t$ & 0.500 & 0.499 & 0.500 & 0.485 & 0.500 & 0.500 & 0.500 & 0.500 \\
           & GT-GAN & 0.492 & 0.490 & 0.405 & 0.315 & 0.490 & 0.500 & 0.390 & 0.335 \\
           & KoVAE & 0.222 & 0.441 & 0.460 & 0.207 & 0.193 & 0.433 & 0.448 & 0.206 \\
           & ImagenI2R & 0.133 & 0.262 & 0.233 & 0.121 & 0.078 & 0.488 & 0.499 & 0.378 \\
           & Ours & \cellcolor{blue!10}$\boldsymbol{0.019}$ & \cellcolor{blue!10}$\boldsymbol{0.025}$ & \cellcolor{blue!10}$\boldsymbol{0.180}$ & \cellcolor{blue!10}$\boldsymbol{0.006}$ & \cellcolor{blue!10}$\boldsymbol{0.019}$ & \cellcolor{blue!10}$\boldsymbol{0.021}$ & \cellcolor{blue!10}$\boldsymbol{0.071}$ &
  \cellcolor{blue!10}$\boldsymbol{0.006}$ \\
          \midrule
          \multirow{5}{*}{\textbf{Pred.}}
           & TimeGAN-$\Delta t$ & 0.210 & 0.470 & 0.450 & 0.115 & 0.300 & 0.490 & 0.600 & 0.080 \\
\bottomrule
      \end{tabular}
  \end{table}

\clearpage

\clearpage
\section*{NeurIPS Paper Checklist}

\begin{enumerate}

\item {\bf Claims}
    \item[] Question: Do the main claims made in the abstract and introduction accurately reflect the paper's contributions and scope?
    \item[] Answer: \answerYes{}
    \item[] Justification: The abstract and introduction (\cref{sec:intro}) clearly state the closed-loop MCEM framework and the \pslr{} E-step recipe, the four quantitative improvements (discriminative, predictive, Context-FID, feature-correlation), and the controlled illustration; each claim is substantiated in \cref{sec:method,sec:results} and the per-dataset tables of \cref{app:additional-exp}.

\item {\bf Limitations}
    \item[] Question: Does the paper discuss the limitations of the work performed by the authors?
    \item[] Answer: \answerYes{}
    \item[] Justification: \cref{sec:conclusions} discusses three limitations: identifiability under heavy corruption, dependence on initialisation due to local optima of MCEM, and wall-clock cost ($\sim\!2.2\times$ ImagenI2R, \cref{tab:wallclock-comparison}).

\item {\bf Theory assumptions and proofs}
    \item[] Question: For each theoretical result, does the paper provide the full set of assumptions and a complete (and correct) proof?
    \item[] Answer: \answerYes{}
    \item[] Justification: Proposition~\ref{prop:frozen-bias} is stated with assumptions and proven in \cref{app:why-close-full}; Neal--Hinton MCEM monotonicity is reviewed and instantiated in \cref{app:em-convergence}; the three structural misspecifications (M1--M3) of single-space samplers in the dual-space setting are derived in \cref{app:mmps-fails} and pinned to the corresponding \pslr{} corrections (C1--C3) in \cref{subsec:pslr}.

    \item {\bf Experimental result reproducibility}
    \item[] Question: Does the paper fully disclose all the information needed to reproduce the main experimental results of the paper to the extent that it affects the main claims and/or conclusions of the paper (regardless of whether the code and data are provided or not)?
    \item[] Answer: \answerYes{}
    \item[] Justification: The paper provides extensive implementation details including training parameters, model architecture (\cref{sec:method}, \cref{app:algorithms}, and the hyperparameter tables \cref{tab:hp_seq_24,tab:hp_seq_96,tab:hp_seq_768} in the appendix), datasets used (\cref{app:datasets}), evaluation metrics (\cref{app:metrics}), and ablation studies (\cref{sec:ablation}).

\item {\bf Open access to data and code}
    \item[] Question: Does the paper provide open access to the data and code, with sufficient instructions to faithfully reproduce the main experimental results, as described in supplemental material?
    \item[] Answer: \answerYes{}
    \item[] Justification: Code is available at \url{https://github.com/azencot-group/PSLR}. Its README gives the environment setup, a download link for all datasets, and the commands to run the method. All datasets are public (\cref{app:datasets}), and the algorithms and hyperparameters are given in \cref{app:algorithms,app:hyperparams}.

\item {\bf Experimental setting/details}
    \item[] Question: Does the paper specify all the training and test details (e.g., data splits, hyperparameters, how they were chosen, type of optimizer) necessary to understand the results?
    \item[] Answer: \answerYes{}
    \item[] Justification: Training/test splits, optimizer settings, and per-dataset hyperparameters are provided in \cref{sec:results} and in \cref{app:hyperparams,app:datasets}, with the global algorithmic knobs ($c$, $n_{\mathrm{CG}}$, $K$, $p_{\mathrm{proj}}$, STL) specified in \cref{app:algorithms} and architectural settings tabulated in \cref{tab:hp_seq_24,tab:hp_seq_96,tab:hp_seq_768}.

\item {\bf Experiment statistical significance}
    \item[] Question: Does the paper report error bars suitably and correctly defined or other appropriate information about the statistical significance of the experiments?
    \item[] Answer: \answerYes{}
    \item[] Justification: Real-data metrics are averaged over $10$ independent runs across three missing rates ($30\%$, $50\%$, $70\%$) (\cref{sec:results}), and the controlled illustration aggregates over $21$ configurations and reports both medians and per-pair gap envelopes ($\Delta_{\mathrm{disc}}$ minimum / median / maximum) in \cref{app:illustration-robustness}, together with one-sigma scatter on the EM trajectory (e.g., $0.029 \pm 0.003$).

\item {\bf Experiments compute resources}
    \item[] Question: For each experiment, does the paper provide sufficient information on the computer resources (type of compute workers, memory, time of execution) needed to reproduce the experiments?
    \item[] Answer: \answerYes{}
    \item[] Justification: All wall-clock measurements were taken on a single NVIDIA RTX~3090 GPU (see the Methodology paragraph of \cref{app:convergence}). Per-iteration training cost is tabulated in \cref{tab:hp_seq_24,tab:hp_seq_96,tab:hp_seq_768} ($40$ M-step epochs/iter; batch size $16$--$128$ set per dataset and sequence length), and end-to-end cost is reported in \cref{tab:wallclock-comparison}, averaged across ten datasets and three missing rates ($30\%/50\%/70\%$).

\item {\bf Code of ethics}
    \item[] Question: Does the research conducted in the paper conform, in every respect, with the NeurIPS Code of Ethics \url{https://neurips.cc/public/EthicsGuidelines}?
    \item[] Answer: \answerYes{}
    \item[] Justification: The paper does not raise ethical concerns; all datasets are public, and the work does not involve sensitive data, scraped content, or human participants.

\item {\bf Broader impacts}
    \item[] Question: Does the paper discuss both potential positive societal impacts and negative societal impacts of the work performed?
    \item[] Answer: \answerYes{}
    \item[] Justification: The paper highlights potential positive applications in healthcare, finance, and the sciences (Abstract and \cref{sec:intro}, including privacy-preserving proxies, stress testing, and dataset balancing), and discusses risks tied to ambiguous data imputation and overreliance on imputer-induced biases (\cref{sec:method}).

\item {\bf Safeguards}
    \item[] Question: Does the paper describe safeguards that have been put in place for responsible release of data or models that have a high risk for misuse (e.g., pre-trained language models, image generators, or scraped datasets)?
    \item[] Answer: \answerNA{}
    \item[] Justification: The paper does not release a general-purpose model or scraped dataset and poses no high risk for misuse.

\item {\bf Licenses for existing assets}
    \item[] Question: Are the creators or original owners of assets (e.g., code, data, models), used in the paper, properly credited and are the license and terms of use explicitly mentioned and properly respected?
    \item[] Answer: \answerYes{}
    \item[] Justification: All datasets used are public and cited correctly (Sine~\citep{yoon2019timegan}, Stocks, MuJoCo~\citep{todorov2012mujoco}, Energy, Weather, ETT-h/m, Electricity); the codebases we build on are academic open-source projects (ImagenTime~\citep{naiman2024imagentime}, ImagenI2R~\citep{fadlon2025imageni2r}, KoVAE~\citep{naiman2024kovae}, GT-GAN~\citep{jeon2022gtgan}, MMPS~\citep{rozet2024diem}, TMPD~\citep{boys2023tmpd}, EDM~\citep{karras2022edm}).

\item {\bf New assets}
    \item[] Question: Are new assets introduced in the paper well documented and is the documentation provided alongside the assets?
    \item[] Answer: \answerYes{}
    \item[] Justification: The new asset is our code, released at \url{https://github.com/azencot-group/PSLR} under the MIT license. Its README documents installation, data preparation, and usage. No new dataset is introduced.

\item {\bf Crowdsourcing and research with human subjects}
    \item[] Question: For crowdsourcing experiments and research with human subjects, does the paper include the full text of instructions given to participants and screenshots, if applicable, as well as details about compensation (if any)?
    \item[] Answer: \answerNA{}
    \item[] Justification: The paper does not involve crowdsourcing or human subjects.

\item {\bf Institutional review board (IRB) approvals or equivalent for research with human subjects}
    \item[] Question: Does the paper describe potential risks incurred by study participants, whether such risks were disclosed to the subjects, and whether Institutional Review Board (IRB) approvals (or an equivalent approval/review based on the requirements of your country or institution) were obtained?
    \item[] Answer: \answerNA{}
    \item[] Justification: The paper does not involve human participants and does not require IRB approval.

\item {\bf Declaration of LLM usage}
    \item[] Question: Does the paper describe the usage of LLMs if it is an important, original, or non-standard component of the core methods in this research? Note that if the LLM is used only for writing, editing, or formatting purposes and does \emph{not} impact the core methodology, scientific rigor, or originality of the research, declaration is not required.
    \item[] Answer: \answerNA{}
    \item[] Justification: LLMs were not used in the methodology, only lightly for text editing and formatting.

\end{enumerate}

\end{document}